\documentclass[twocolumn]{bmcart}
\usepackage{amsthm,amsmath}
\usepackage[utf8]{inputenc} %unicode support
\usepackage{enumitem}
\usepackage{longtable}
\usepackage{multicol}
\usepackage{multirow}
\usepackage[T1]{fontenc}
\usepackage{csquotes}
\usepackage{longtable}
\usepackage{booktabs}
\usepackage{wrapfig}
\usepackage{enumitem}
\usepackage{paralist}
\usepackage{array,multirow,graphicx,adjustbox}
\usepackage{amsfonts}
\usepackage{xcolor}
\usepackage{tabularx}
\usepackage{makecell}
\usepackage{pbox}
\usepackage{subfigure}
\usepackage{natbib}
\usepackage[hang,flushmargin]{footmisc}
\usepackage{flushend}
\usepackage{multicol, lipsum}
\usepackage{pdflscape}
\usepackage{graphicx}
\usepackage{hyperref}
\usepackage{float}

\startlocaldefs
\definecolor{darkgreen}{RGB}{0, 100, 0}
\definecolor{linkcol}{RGB}{0,70,25}
\definecolor{citecol}{RGB}{0,70,25}
\definecolor{urlcol}{RGB}{0,70,25}
\definecolor{celestialblue}{RGB}{74,150,210}
\definecolor{darkblue}{rgb}{0,0,0.5}
\definecolor{darkred}{rgb}{0.6, 0, 0} 

\newcommand{\descr}[1]{\smallskip\noindent\textbf{#1}}

\newlist{mycompactitem}{itemize}{1}
\setlist[mycompactitem,1]{
  label=--,
  left=1pt,
  labelindent=0em,
  itemindent=0em,
  parsep=0pt,
}
\endlocaldefs

\begin{document}
%\onecolumn
%%% Start of article front matter
\begin{frontmatter}

\begin{fmbox}
\dochead{Systematic Review*}

\title{Application of AI-based Models for Online Fraud Detection and Analysis}

%%%%%%%%%%%%%%%%%%%%%%%%%%%%%%%%%%%%%%%%%%%%%%
%%                                          %%
%% Enter the authors here                   %%
%%                                          %%
%% Specify information, if available,       %%
%% in the form:                             %%
%%   <key>={<id1>,<id2>}                    %%
%%   <key>=                                 %%
%% Comment or delete the keys which are     %%
%% not used. Repeat \author command as much %%
%% as required.                             %%
%%                                          %%
%%%%%%%%%%%%%%%%%%%%%%%%%%%%%%%%%%%%%%%%%%%%%%

% \author[
%   addressref={anon},                   % id's of addresses, e.g. {aff1,aff2}
% %  corref={aff1},                       % id of corresponding address, if any
% % noteref={n1},                        % id's of article notes, if any
% email={anon@anon}   % email address
% ]{\inits{AP}\fnm{Author} \snm{Anonymous}}

\author[
  addressref={scs},                   % id's of addresses, e.g. {aff1,aff2}
  % corref={ap},                       % id of corresponding address, if any
% noteref={n1},                        % id's of article notes, if any
  email={antonis.papasavva@ucl.ac.uk}   % email address
]{\inits{AP}\fnm{Antonis} \snm{Papasavva}}
\author[
  addressref={scs},                   % id's of addresses, e.g. {aff1,aff2}
  % corref={sj},                       % id of corresponding address, if any
% noteref={n1},                        % id's of article notes, if any
  email={shane.johnson@ucl.ac.uk}   % email address
]{\inits{SJ}\fnm{Shane} \snm{Johnson}}
\author[
  addressref={arc},                   % id's of addresses, e.g. {aff1,aff2}
  % corref={ed},                       % id of corresponding address, if any
% noteref={n1},                        % id's of article notes, if any
  email={e.lowther@ucl.ac.uk}   % email address
]{\inits{EL}\fnm{Ed} \snm{Lowther}}
\author[
  addressref={aru},                   % id's of addresses, e.g. {aff1,aff2}
  % corref={sl},                       % id of corresponding address, if any
% noteref={n1},                        % id's of article notes, if any
  email={samantha.lundrigan@aru.ac.uk}   % email address
]{\inits{SL}\fnm{Samantha} \snm{Lundrigan}}
\author[
  addressref={scs},                   % id's of addresses, e.g. {aff1,aff2}
  % corref={em},                       % id of corresponding address, if any
% noteref={n1},                        % id's of article notes, if any
  email={e.mariconti@ucl.ac.uk}   % email address
]{\inits{EM}\fnm{Enrico} \snm{Mariconti}}
\author[
  addressref={aru},                   % id's of addresses, e.g. {aff1,aff2}
  % corref={am},                       % id of corresponding address, if any
% noteref={n1},                        % id's of article notes, if any
  email={anna.markovska@aru.ac.uk}   % email address
]{\inits{AM}\fnm{Anna} \snm{Markovska}}
\author[
  addressref={scs},                   % id's of addresses, e.g. {aff1,aff2}
  %corref={nt},                       % id of corresponding address, if any
% noteref={n1},                        % id's of article notes, if any
  email={n.tuptuk@ucl.ac.uk}   % email address
]{\inits{NT}\fnm{Nilufer} \snm{Tuptuk}}

%%%%%%%%%%%%%%%%%%%%%%%%%%%%%%%%%%%%%%%%%%%%%%
%%                                          %%
%% Enter the authors' addresses here        %%
%%                                          %%
%% Repeat \address commands as much as      %%
%% required.                                %%
%%                                          %%
%%%%%%%%%%%%%%%%%%%%%%%%%%%%%%%%%%%%%%%%%%%%%%

% \address[id=anon]{%                           % unique id
%  \orgdiv{Department},             % department, if any
%  \orgname{Anon Uni},          % university, etc
%  \city{City},                              % city
%  \cny{country}                                    % country
% }

\address[id=scs]{%                           % unique id
  \orgdiv{Security and Crime Science},             % department, if any
  \orgname{University College London},          % university, etc
  \city{London},                              % city
  \cny{United Kingdom}                                    % country
}

\address[id=aru]{%                           % unique id
  \orgdiv{Humanities and Social Sciences},             % department, if any
  \orgname{Anglia Ruskin University},          % university, etc
  \city{London},                              % city
  \cny{United Kingdom}                                    % country
}

\address[id=arc]{%                           % unique id
  \orgdiv{Advanced Research Computing Centre},             % department, if any
  \orgname{University College London},          % university, etc
  \city{London},                              % city
  \cny{United Kingdom}                                    % country
}

\textbf{\textcolor{darkred}{This manuscript has been accepted in the Crime Science Journal. Please cite accordingly. \\ DOI: \href{dx.doi.org/10.1186/s40163-025-00248-8}{10.1186/s40163-025-00248-8}}}

%%%%%%%%%%%%%%%%%%%%%%%%%%%%%%%%%%%%%%%%%%%%%%
%%                                          %%
%% Enter short notes here                   %%
%%                                          %%
%% Short notes will be after addresses      %%
%% on first page.                           %%
%%                                          %%
%%%%%%%%%%%%%%%%%%%%%%%%%%%%%%%%%%%%%%%%%%%%%%

%\note{Sample of title note}     % note to the article
%\note[id=n1]{Equal contributor} % note, connected to author

%\end{fmbox}% comment this for two column layout

%%%%%%%%%%%%%%%%%%%%%%%%%%%%%%%%%%%%%%%%%%%%%%
%%                                          %%
%% The Abstract begins here                 %%
%%                                          %%
%% Please refer to the Instructions for     %%
%% authors on http://www.biomedcentral.com  %%
%% and include the section headings         %%
%% accordingly for your article type.       %%
%%                                          %%
%%%%%%%%%%%%%%%%%%%%%%%%%%%%%%%%%%%%%%%%%%%%%%

\begin{abstractbox}

\begin{abstract} % abstract
\parttitle{Background} %if any
Fraud is a prevalent offence that extends beyond financial loss, impacting victims emotionally, psychologically, and physically. 
Advances in online communication technologies continue to create new opportunities for fraud, and fraudsters increasingly using these channels for deception. With the progression of technologies like Generative Artificial Intelligence (GenAI), there is a growing concern that fraud will increase in scale using these advanced methods, with offenders employing deep-fakes in phishing campaigns, for example. However, the application of AI to detect and analyse patterns of online fraud remains understudied. This review addresses this gap by investigating the potential role of AI in analysing online fraud using text data.

\parttitle{Methods} %if any
We conducted a Systematic Literature Review (SLR) to investigate the application of AI and Natural Language Processing (NLP) techniques for online fraud detection.
The review adhered to the PRISMA-ScR protocol, with eligibility criteria including language, publication type, relevance to online fraud, use of text data, and AI methodologies.
Out of $2,457$ academic records screened, $350$ met our eligibility criteria, and $223$ were analyzed and included herein.

\parttitle{Results}
We discuss the state-of-the-art NLP techniques used to analyse various online fraud categories; the data sources used for training the NLP models; the NLP algorithms and models built; and the performance metrics employed for model evaluation. 
We find that the current state of research on online fraud is broken into the various scam activities that take place, and more specifically, we identify $16$ different frauds that researchers focus on.
Finally, we present the most recent and best-performing AI methods employed for detecting online scams and fraud activities.

\parttitle{Conclusions}
This SLR enhances academic understanding of AI-based detection methods for online fraud and offers insights for policymakers, law enforcement, and businesses on safeguarding against such activities. We conclude that existing approaches focusing on specific scams are unlikely to generalise effectively, as they will require new models to be developed for each fraud type.

Furthermore, we conclude that the evolving nature of scams limits the effectiveness of models trained on outdated data. 
We also identify that researchers often omit discussions of the limitations of their data or training biases.
Finally, we find issues in the consistency with which the performance of models is reported, with some studies selectively presenting metrics, leading to potential biases in model evaluation.
% Specifically, a model can perform well on its testing data, but this can lead to \emph{overfitting}, which fails to generalize to new, unseen data. 
% Additionally, we highlight issues in performance reporting across various studies, noting that many studies selectively report certain performance metrics, leading to biased assessments of model performance.
\end{abstract}

%%%%%%%%%%%%%%%%%%%%%%%%%%%%%%%%%%%%%%%%%%%%%%
%%                                          %%
%% The keywords begin here                  %%
%%                                          %%
%% Put each keyword in separate \kwd{}.     %%
%%                                          %%
%%%%%%%%%%%%%%%%%%%%%%%%%%%%%%%%%%%%%%%%%%%%%%


% MSC classifications codes, if any
%\begin{keyword}[class=AMS]
%\kwd[Primary ]{}
%\kwd{}
%\kwd[; secondary ]{}
%\end{keyword}

\end{abstractbox}
\end{fmbox}% uncomment this for twcolumn layout

\end{frontmatter}
%\twocolumn

%%%%%%%%%%%%%%%%%%%%%%%%%%%%%%%%%%%%%%%%%%%%%%
%%                                          %%
%% The Main Body begins here                %%
%%                                          %%
%% Please refer to the instructions for     %%
%% authors on:                              %%
%% http://www.biomedcentral.com/info/authors%%
%% and include the section headings         %%
%% accordingly for your article type.       %%
%%                                          %%
%% See the Results and Discussion section   %%
%% for details on how to create sub-sections%%
%%                                          %%
%% use \cite{...} to cite references        %%
%%  \cite{koon} and                         %%
%%  \cite{oreg,khar,zvai,xjon,schn,pond}    %%
%%  \nocite{smith,marg,hunn,advi,koha,mouse}%%
%%                                          %%
%%%%%%%%%%%%%%%%%%%%%%%%%%%%%%%%%%%%%%%%%%%%%%

%%%%%%%%%%%%%%%%%%%%%%%%% start of article main body
% <put your article body there>

%%%%%%%%%%%%%%%%
%% Background %%
%%
\section{Introduction}\label{sec:intro}
Online fraud has emerged as one of the most pervasive and challenging threats in the digital age, affecting individuals of all ages, businesses of different sizes, and governments. 
Defined broadly, online fraud is an umbrella term that involves acts of deception or deliberate impersonation on the Internet for the personal gain of the fraudster, often resulting in a financial loss for the victim \cite{ukfinance2023}. 
In addition to financial losses, fraud can have a wide range of impacts on victims. 
These include emotional and psychological effects such as anger, fear, shame, depression, loss of confidence, and trauma; impacts on physical and mental well-being; it can harm relationships and lead to loneliness and isolation; and cause negative changes in behaviour~\cite{parliament_post_2020}. 
Although evidence suggests that certain sociodemographic groups face higher risks of fraud (e.g., women aged 25-44 and those in the highest income bracket), fraud affects individuals across all demographics~\cite{parliament_post_2020} and sometimes in different ways. 
For example, in a UK study, victims earning £20,000 or less, those aged 65 and over, and female victims reported that fraud impacted their self-confidence more than did victims in general.

For the year ending March 2023, the Crime Survey for England and Wales estimated that 3.5 million fraud offences, including online fraud, took place that year~\cite{ons2023}. 
In that year, compared to the year ending March 2020, advance fee fraud increased significantly, from $60,000$ to $391,000$ offences. 
This increase is largely due to society's growing reliance on the Internet and digital platforms for everyday services, transactions, and communications. 
According to The Office of Communications (Ofcom)~\cite{ofcom2023}, $92\%$ of adults in the UK use the Internet for a wide variety of activities, including communication, education, and entertainment. 
Activities such as banking, shopping, and socialising are increasingly happening via online platforms, expanding the landscape for fraudsters to exploit vulnerabilities or use these platforms to deceive victims. 
In 2020, online shopping scams made up $38\%$ of all reported scams worldwide, an increase of $6\%$ compared to the pre-Covid-19 outbreak~\cite{statista-ecommerce-fraud}. 

Online fraud encompasses a wide range of deceptive activities, including identity theft, phishing, advance fee fraud, romance scams, fraudulent investment scams, and more. 
It is important to highlight that there is no universally accepted definition of ``online fraud,'' and the term is often used interchangeably with the term ``scam''. 
Legally, ``fraud is defined as false representation to cause loss to another or to expose another to a risk of loss''~\cite{ukfraudlegislation2006}, and scam is the process where criminals gain the trust of victims to deceive or cheat them~\cite{ukscamglossary2006} through false representation and other means, so that the victim trusts them, which in turn results in various kinds of losses. 

The National Fraud Authority of UK published a literature review~\cite{fraudtypologiesuk}, in which they compared the distinction of the term fraud as defined by the amended Fraud Act 2006~\cite{ukfraudlegislation2006} and the typology produced by Levi~\cite{levi2008definingfraud}. 
They found that fraud embraces a broad scope of crimes, whereas scams often focus on fraud against individuals and small firms. 
For example, different scams like advance fee, romance, tech support, etc., all fall under the fraud umbrella, but they are also deception methods, which are, in part, scams. 
Hence, in this work, we use both terms as various scams represent the different deception methods scammers use to trick victims, while the term fraud includes all scams.

Online frauds exploit the virtual nature of the Internet and the anonymity it provides to reach victims. 
This virtual environment, coupled with jurisdictional challenges (where offenders and victims may be in different regions of the world), makes fraud difficult to detect and prevent using traditional policing techniques. 
The complexity of online fraud is further heightened by its evolving nature, as fraudsters continuously adapt their techniques to bypass new security measures and exploit emerging technologies to target new victims~\cite{skidmore24}. 

Given the scale and impact of online fraud, there is a need for new methods to detect and prevent such activities. 
The use of Natural Language Processing (NLP), in combination with other Artificial Intelligence (AI), has been proposed for identifying, characterising, and detecting fraudulent patterns in applications like phishing~\cite{9795286}, fake job advertisement~\cite{amaar2022detection}, and for the purposes of analysing scam patterns~\cite{lwin2023supporting} which could help develop preventive measures and mitigate risks of online fraud. 
However, understanding the current state of AI techniques in combating online fraud, the data sources used, the evaluation methods for AI models, and the specific types of fraud that are most prevalent, remains a significant challenge. 
This is due to the constant emergence of new fraud activities that use various communication mediums and social engineering attacks, in an attempt by fraudsters to remain undetected. 
Therefore, there is a pressing need to shift from detecting and analysing the effects of fraud to the early detection of emerging fraudulent activities online and new methods of social engineering.

This study aims to address these challenges by conducting a comprehensive review of the state-of-the-art AI techniques used to detect fraudulent online activities. 
Specifically, we examine the data sources widely used by researchers to study online fraud, the methods researchers use to evaluate the developed AI models, and the most popular types of online fraud targeted in their studies. By synthesising findings from academic papers, this review aims to provide a thorough understanding of the current landscape of online fraud detection and prevention, highlighting gaps in existing research, and proposing directions for future studies.

\descr{Manuscript Structure.} The rest of the paper is organised as follows. The next section (\S\ref{sec:online-fraud}) introduces various well-known types of online fraud and provides a detailed discussion of the latest and most widely used AI methodologies, including how they are evaluated. 
The background review conducted for this section helped us to formulate and refine our research questions.

Section~\ref{sec:systematic-review} outlines the methodology followed in this SLR, including the protocol, the criteria used to filter eligible papers, and the data extracted from each study. 
The results of the SLR are then presented in Section~\ref{sec:slr_results}.
In Section~\ref{sec:summary-of-findings}, we discuss the findings of our literature review, categorized by the various types of online fraud identified, and provide detailed insights into how each of our research questions were addressed.

Finally, Section~\ref{sec:discussion} offers a deeper analysis of our findings, highlighting limitations and shortcomings in the reporting of AI models, particularly regarding performance and data sources. 
We also propose recommendations for researchers developing detection models for online fraud, before concluding in Section~\ref{sec:conclusion}.

\section{Online Fraud and AI}\label{sec:online-fraud}
Online fraud refers to any deliberate act of deception conducted over the Internet to cause an unlawful or unfair loss~\cite{ukfraudlegislation2006}.
It involves exploiting online platforms, services, and technologies to deceive individuals or organisations for financial, personal, or material gain. 
Online fraud can take many forms, each characterised by the method of deception and the medium used.

\subsection{Fraud Categories}
The list of online fraud activities is extensive and constantly evolving, with new types and sub-types emerging~\cite{skidmore24}.
To conduct our SLR, it was important first to identify the most prevalent types of offences likely to be analysed by the studies included.
To briefly discuss online fraud, we studied various taxonomies, studies, and reports published or discussed by UK government bodies~\cite{fraudtypologiesuk,parliament_post_2020}, financial services~\cite{ukfinance2023}, telecommunication providers~\cite{ofcom_report}, policing think tanks~\cite{skidmore24}, and academics~\cite{rabitti2024taxonomy,zhou2024understanding,levi2008definingfraud}.

Developing a comprehensive taxonomy or classification for all online fraud activities requires special attention, which is beyond the scope of this work. 
Note that many taxonomies, especially the ones published by UK government bodies~\cite{fraudtypologiesuk,parliament_post_2020} also discuss fraud and crime that potentially can take place offline, which we omit in the following discussion of online fraud as offline crime falls beyond the focus of our SLR.
Below, we outline some of the well-known and most discussed online fraud types we encountered while performing our preliminary research on online fraud, aided by the reports discussed above.
We note that the following is not intended as a complete taxonomy of online fraud, nor does it represent the findings of this SLR. Instead, it is intended to briefly discuss popular online frauds and scams for the reader.

\begin{mycompactitem}
    \item \textbf{Phishing} is the process where fraudsters impersonate representatives of legitimate organisations or acquaintances of the targeted victim to trick them into providing personal information such as usernames, passwords, credit card details, or bank account details. This activity can be done through various mediums, like email, phone calls (aka Vishing), SMS (aka Smishing), and any other way of online communication. Various phishing scams have surfaced over the years, including the Royal Mail scam ~\cite{royalmailscams}, banking scams~\cite{barcleysscams}, and HMRC scams~\cite{hmrcscams}. Notably, phishing scams often include deceptive web addresses created by cybercriminals to trick victims into believing they are visiting legitimate websites. The primary goal of these URLs is to steal personal data, including usernames, passwords and credit card details, for financial gain.
    \item \textbf{Fake Reviews} are deceptive or fraudulent reviews created to mislead potential customers about the quality, reliability, or legitimacy of a product, service, or app. On fraudulent e-commerce websites and app stores, fake reviews play a crucial role in tricking victims into trusting and using fraudulent apps or purchasing substandard or non-existent products. This leads to potential victims trusting fraudulent websites, services, or apps, providing them with their credit card details for a purchase, which leads the victim to a vulnerable position~\cite{paul2021fake}.
    \item \textbf{Recruitment Fraud} is a type of online scam where fraudsters pose as legitimate employers or recruiters to deceive job seekers. The primary goal of these scams is to receive ``fees'' for a job application, steal personal information, extort money, or exploit the victim in some other way. This type of fraud preys on individuals seeking employment, often targeting those who are most vulnerable or desperate for work~\cite{mehboob2021smart}.
    \item \textbf{Romance Fraud} (aka romance scams or dating scams) involves fraudsters creating fake profiles on dating websites, social media, or other online platforms to deceive victims into believing they are in a genuine romantic relationship. The primary objective is to exploit the victim's emotions to extort money, personal information, or other benefits. This elaborate scam is extremely difficult to detect since it is also under-reported due to victims feeling ashamed and hurt for being victimised by someone they considered to be a romantic partner~\cite{coluccia2020online}. In these scams, fraudsters communicate with victims for a long time before presenting them with an ``investment opportunity'' or requesting their financial aid. Romance scams are closely related to \emph{Cryptocurrency Pig Butchering scams}~\cite{cross2023romance}, where victims are gradually lured into making increasing contributions over a long period of time, usually in cryptocurrency, to a fraudulent scheme~\cite{ordekian2024sinister}.
    \item \textbf{Fraudulent Investment} includes scams where fraudsters promise victims significant winnings or lucrative opportunities~\cite{vasek2015there}. These scams are usually associated with the romance scams discussed above. Once the victims try to withdraw their ``winnings,'' the scammers will extort them by asking for ``fees'' and ``taxes'' to be paid in advance. The promised benefits and winnings never materialize, and the initial investment sums and fees are lost~\cite{agarwal2023defi}. Fraudulent investment is the umbrella that covers Cryptocurrency Pig Butchering scams explained above, and various \emph{Ponzi schemes}~\cite{vasek2019analyzing} where early investors greatly benefit from the investments of later investors, also known as \emph{pyramid schemes}.
    \item \textbf{Crypto Market Manipulation} involves artificially increasing or decreasing the price of cryptocurrencies to achieve financial gain. It often involves coordinated efforts by individuals or groups to manipulate the market to create false perceptions of supply, demand, or market sentiment. Some common techniques used in crypto market manipulation include: \emph{Pump and Dump}, which inflates the price of a cryptocurrency through misleading or false statements (pumping), encouraging others to buy, and then selling off the cryptocurrency at a profit once the price has been pumped up (dumping); \emph{Wash Trading} occurs when a trader buys and sells the same cryptocurrency simultaneously to create deceptive activity on the market; \emph{Spoofing} involves placing significant buy or sell orders to withdraw them before execution to mislead perceptions related to the market demand or supply; \emph{Front-running} involves placing orders ahead of a large trade that is known to occur, to benefit from the subsequent price movement caused by the large trade; and many others~\cite{hamrick2021examination}. These scams are also similar to \emph{Stock Market Manipulation}.
    \item \textbf{Fraudulent E-Commerce} involves deceptive practices or scams conducted through online e-commerce platforms. These scams aim to exploit digital payment systems to deceive consumers or businesses by paying for a fraudulent product or service.
    \item \textbf{Fraudulent Crowdfunding} refers to the misuse of crowdfunding platforms to deceive donors or backers, often by providing false or misleading information about a crowdfunding campaign's nature, purpose, or outcome. Crowdfunding is a method of raising money from many people via online platforms to fund projects, products, or causes~\cite{cumming2021disentangling}. A fraud similar to crowdfunding is \emph{Charity Fraud} and \emph{Disaster Scams}, where scammers seek donations for organisations that do not exist or do little work. These scams are particularly common after high-profile disasters as criminals often use tragedies to exploit people who are looking to donate~\cite{charityfraud}.
    \item \textbf{Gambling Fraud} is any illegal activity that is intended to cheat players or an online gambling platform. Fraudsters manage to trick victims and platforms in different ways, including rigged games, fake websites (phishing URLs described above), account takeovers (via stealing legitimate users' access codes), and creating fake apps with fake reviews, as discussed above, to gain the trust of users. Online gambling fraud can happen on multiple platforms and involve a wide variety of games, including \emph{casino scams, sports betting scams}, and \emph{lottery scams}~\cite{hong2022analyzing}.
    \item \textbf{Tax Scams} occur when scammers falsify information regarding pending tax money or maliciously impersonate tax officials to trick individuals or business entities into wilfully paying them ``fees''~\cite{brody2014income}. Scams similar to tax scams are \emph{Council Tax Scams}, various \emph{Utility bill scams}, \emph{Insurance Scams}, etc. These scams fall under the umbrella of \emph{Phishing} as they often take place via SMS, phone calls, or emails.
    \item \textbf{Pension Scams} are similar to Tax Scams. Scammers aim to make money through fees, direct access to pension savings, or by receiving investments~\cite{pensionscams}.
\end{mycompactitem}

The complexity and interconnected nature of scams and frauds make categorising them under a single typology challenging~\cite{skidmore24,cohen2019investigation}. 
Phishing scams, for instance, serve as a broad umbrella that covers phishing conducted using various methods like vishing (voice) and smishing (SMS). 
Yet, they can also be integral parts of investment scams when scammers develop phishing websites to gain victims' trust. 
Similarly, most scams often involve the scammer impersonating an authority (government, law enforcement), friend, organisation (e.g., bank), or other entity (e.g., delivery service), making impersonation scams difficult to break down as they are an integral part of other scams (e.g., a delivery scam is also an impersonation scam as the offender is clearly not an Amazon representative, for example).

To summarise, different scam types frequently overlap, blurring the lines between distinct categories and demonstrating today's intricate web of fraudulent activities. 
The multifaceted nature of these scams highlights the difficulty in creating a comprehensive classification system that can effectively encompass all types of fraudulent schemes.
We discuss this challenge and limitation in detail before concluding our SLR, in our Discussion (Section~\ref{sec:discussion}).

\subsection{AI techniques}
This study investigates AI-based techniques for processing unstructured text data to analyse fraud. 
Much of this text data, like news articles, research papers, government reports, books, social media posts (tweets and Facebook comments), communications (such as emails, SMS messages, and chat logs), and web content (reviews on online marketplaces, travel and hospitality platforms, and comments on video sharing platforms), is inherently unstructured. 
Statista~\cite{Statista} estimated that the global open data that is accessible on the entire Web was 64.2 zettabytes in 2020, and it is expected to exceed 180 zettabytes by 2025.
With each new digital platform or communication channel, this data is increasing. 
Most of the data created is unstructured text that provides opportunities for understanding human behaviour, habits, opinions and experiences. 
It contains information about users’ experiences, events, themes, opinions, and sentiments that can be important for deriving meaningful insight from their experience related to fraudulent activities. 
Manual traditional data analysis techniques, like keyword searches and the coding of themes, are often limited, and the extraction of meaningful insights at scale is unachievable, making advanced computer-driven automated techniques necessary. 

However, there are often significant challenges associated with the analysis of this data due to the diversity of natural (human) language used. 
This includes dealing with noise (irrelevant or useless data), a wide array of linguistic variations of human language due to regional or cultural nuances, the use of slang or jargon, abbreviations, spelling errors, typos and grammatical mistakes, which often pose challenges for the efficient analysis of text data. 
NLP techniques were designed to effectively understand the structure (syntax) and comprehend (semantic) spoken and written human language the way humans do. 
Advancements in AI, including machine and deep learning, along with improvements in technology (such as increased computing power) and software (such as the availability of programming tools and libraries), have significantly improved the ability to process and understand large volumes of unstructured text. 
These tools have been widely used in many fields, from sales and marketing to spam detection. 
The process of collecting, pre-processing, and training AI-based models using text data often involves the pipeline shown in Figure \ref{fig:DataProcess}:

\begin{figure}[t]
\centering
\includegraphics[width=0.97\linewidth]{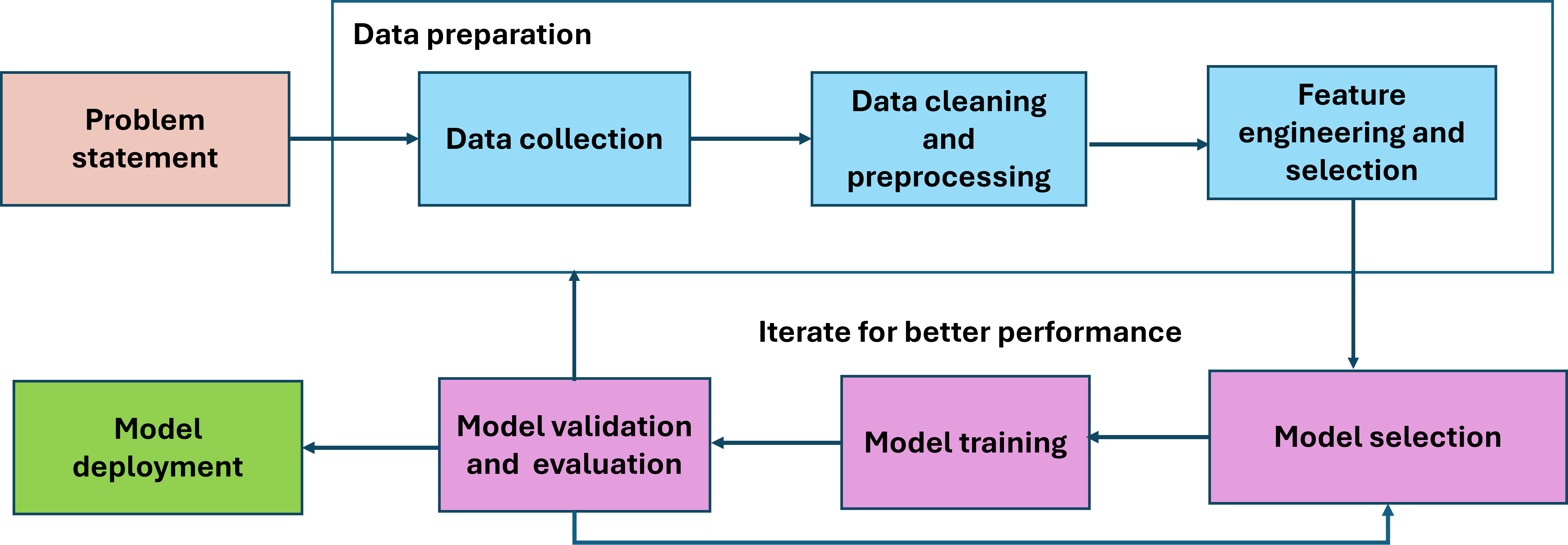}
\caption{Common pipeline for NLP-based models}
\label{fig:DataProcess}
\end{figure}

\begin{mycompactitem}
    \item \textbf{Problem statement}: Using domain knowledge, a suitable research question is formulated for AI to address. This could be a classification problem (e.g. to classify text into a number of categories), or explanatory analysis involving the identification of patterns within a text. 
    \item \textbf{Data preparation}: AI-based models require the collection of appropriate data towards the building of a \textit{corpus} (a collection of structured sets of \textit{documents} such as emails, news articles, social media posts, or transcripts) used to train models to analyse the research question. Often, the acquired data comes as unstructured data and requires cleaning and pre-processing. The data cleaning and pre-processing involve removing unwanted or redundant data to reduce the noise in the data. This may include removing duplicates or incomplete entries, symbols, punctuations, numbers, stop-words, converting acronyms to full words, and handling non-English words, slang, or jargon. Further pre-processing may involve text normalisation techniques like \textit{stemming} or \textit{lemmatisation} to reduce words to their root or base form to improve the accuracy of text analysis. 
    \item \textbf{Feature engineering and selection}: Feature engineering involves preparing data for machine learning models. It consists of extracting and selecting predictive features in supervised learning or finding patterns in unlabeled data in unsupervised learning. This task requires using domain knowledge to develop and select appropriate features. Common text features often used are \textit{n-grams} (sequences of \textit{n} consecutive words); \textit{Term Frequency-Inverse Document Frequency (TF-IDF)} (a statistical method that weights the importance of a word/term in a document within a corpus) matrix; sentiments and emotions present; lexical features (e.g. presence of certain words, Keyword-in-Context and lexical diversity); syntactic features (e.g. Part-of-Speech tags); semantic features (e.g. entities mentioned, word-embedding); readability scores; structural features (e.g. length of the text, number of paragraphs); and domain-specific features (e.g. presence of specialised terms). 
    \item \textbf{AI technique/algorithm selection}: This step involves selecting an appropriate AI algorithm for building the AI-based model. Tasks associated with text often involve two main categories of AI-based models: \textit{supervised} and \textit{unsupervised} machine learning. The choice of the algorithm will depend on the learning, and type of AI required. Supervised machine learning algorithms are often used for text classification problems. The learning algorithm is fed with input features (training data) and labels (discrete outputs). The supervised machine learning algorithm aims to map input features to discreet outputs. Traditional supervised machine learning algorithms include Logistic Regression (LR), Naive Bayes (NB), Decision Trees (DT), Random Forest (RF - multiple DTs), Support Vector Machines (SVM), and K-Nearest Neighbors (KNN). New supervised machine learning techniques include neural networks (NN) and deep learning-based models. Unsupervised machine learning models, often used in exploratory data analysis, involve working with unlabelled data to discover hidden patterns and themes. Unsupervised machine learning algorithms include clustering techniques using algorithms like K-means, hierarchical clustering and Density-Based Spatial Clustering (DBSCAN); and topic modelling, achieved by algorithms like Latent Dirichlet Allocation (LDA) and Latent Semantic Analysis (LSA). 

\begin{figure}[t]
\centering
\includegraphics[width=0.9\linewidth]{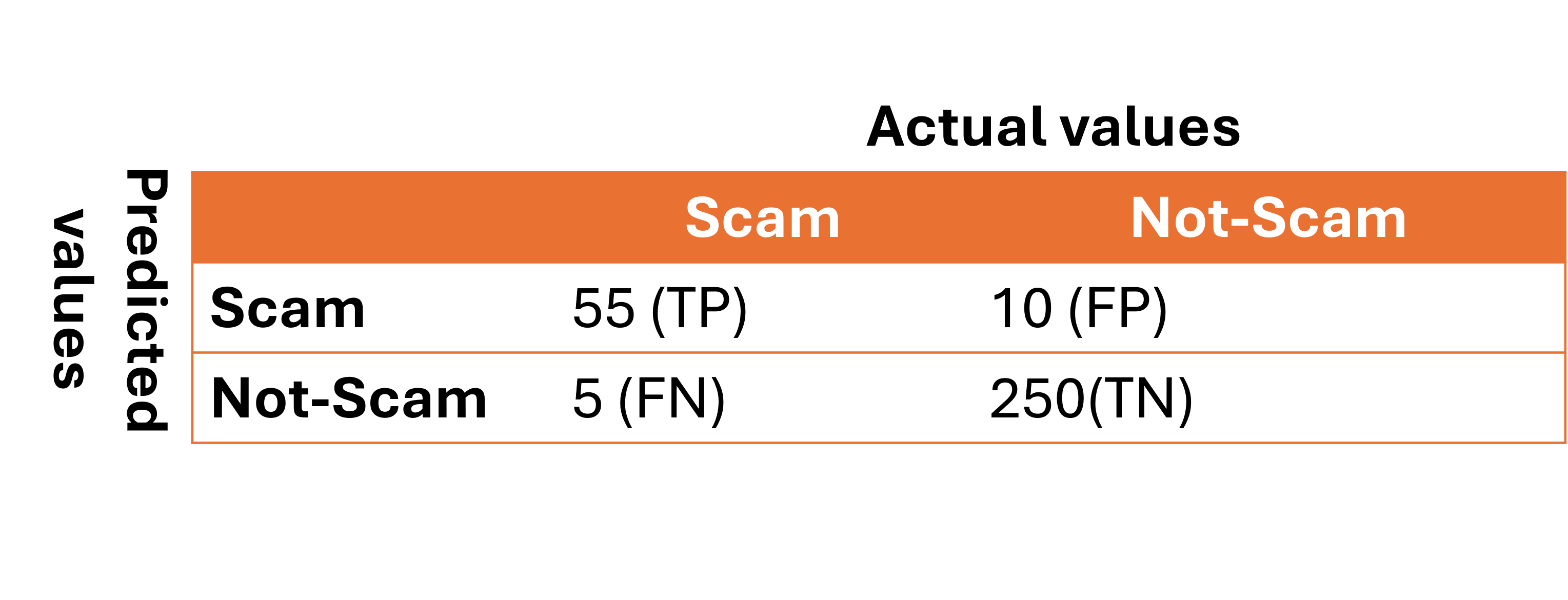}
\caption{Confusion Matrix}
\label{fig:confusionmatrix}
\end{figure}
    
    \item \textbf{Model training}: Often, machine learning algorithms will have parameters that need to be tuned before learning begins, known as hyperparameters. The tuning process involves re-training the model multiple times using different values for these hyperparameters and selecting the best combination of values based on model performance on a metric of interest. In the case of supervised machine learning, the hyperparameters might be tuned using model performance on different “folds” of the data in an approach known as cross-validation. With this approach, a randomly selected proportion of the data is kept separate from the training data, and used for final model evaluation. This approach provides the best indication of how the model will likely perform on new, unseen data. In the case of unsupervised modelling, heuristics are used to identify the optimal number of clusters or topic modelling.  
    \item \textbf{Model evaluation}: The model's performance needs to be evaluated. In the case of supervised modelling, this will involve measuring the model's performance on the test data. The classic supervised machine learning algorithms can be evaluated using performance metrics such as a confusion matrix (Figure \ref{fig:confusionmatrix}), accuracy, precision, recall, F1-score, sensitivity, specificity, Receiver Operating Characteristic (ROC) curve, and the Area Under the Curve (AUC) curve: 
\[Accuracy = \frac{TP+TN}{TP+TN+FP+FN} \]
\[Precision = \frac{TP}{TP+FP} \]
\[Recall = \frac{TP}{TP+FN} \]
\[F1-score = \frac{2*Precision*Recall}{Precision+Recall}\]
Where:\\
\emph{True Positives (TP)}: The model correctly predicts a positive class (e.g., those that were classified as scam.) \\
\emph{True Negatives (TN)}: The model correctly predicts a negative class (e.g., those classified as not-scam). \\
\emph{False Positives (FP)}: The model incorrectly predicts the positive class (e.g. not-scam is predicted as a scam). \\
\emph{False Negatives (FN)}: The model incorrectly predicts the negative class (e.g. scam predicted as not scam).\\
 \\
Sensitivity is the same as recall or the true positive rate, and it captures the model's ability to identify the positive class (i.e. scam cases) correctly: 

\[Sensitivity (TPR) = Recall = \frac{TP}{TP+FN}\]

Specificity, also known as false positive rate (FPR), measures the proportion of true negatives, and it captures the model's ability to identify negative class (i.e. not-scam cases) correctly: 

\[Specificity = \frac{TN}{FP+TN}\]

The ROC curve illustrates the performance of one or more binary classifiers. It plots the sensitivity against the 1-specificity for various thresholds. The AUC is calculated as the area under the ROC curve. 

     \item \textbf{Deploy model}: Once the models work well, they can be deployed. When considering deployment of the model, one must address questions regarding why others should trust the model, how the model arrived at its conclusions and usability, and carefully assess the ethical implications of AI to ensure its suitability for deployment and that it is not biased. 
\end{mycompactitem}

In unsupervised machine learning models, due to a lack of ground truth labels, the performance of the model evaluation may involve subjective interpretation to interpret the outputs (e.g. clusters or topics) generated by the model. 

\subsection{Advanced NLP techniques}
This section briefly introduces advanced NLP techniques, aiming to familiarise the reader with these concepts as they are later referred to during the SLR findings. 

\textit{Word embeddings} is an important technique in NLP that involves encoding words as vectors of real numbers that are designed to capture their similarities. 

Words closer together in the vector space are expected to have similar meanings or relationships. Two of the widely used word embedding techniques are Word2Vec and GloVe. Word2Vec uses a simple neural network trained on large text datasets iteratively to predict either context words or target words. Word2Vec uses two approaches \cite{DBLP:journals/corr/Rong14}: Continuous Bag of Words (CBOW) predicts the target word based on its surrounding context words, whereas Skip-gram predicts surrounding context words based on a given target word. In the sentence `The quick brown fox jumps over the lazy dog', if `fox' is used as the target word, the CBOW model uses `The', `quick', `brown', `jumps', `over', `the', `lazy', and `dog' as context and predicts the word `fox.' In Skip-gram, `fox' is used to predict the surrounding words like `The', `quick', `brown', `jumps', `over', `the', `lazy', and `dog'.
\textit{Global Vectors for Word Representation} (GloVe) \cite{glove} learns the vector representation of words using global word-word co-occurrence statistics obtained from the training data to show the semantic relationships between words. 

Large Language Models use word embeddings to generate responses to natural language inputs. 

\textit{Large Language Models} (LLMs) \cite{zhao2023surveylargelanguagemodels} are advanced NLP tools trained on billions of words from a wide variety of sources and are designed to perform complex tasks like translations, summarisation and the performance of human-like conversational abilities. Most LLMs are developed using a transformer-based architecture (\textit{transformers}) \cite{DBLP:journals/corr/VaswaniSPUJGKP17}, and billions of parameters are used for training. 
Transformers are a type of deep-learning neural network model, and they are more efficient compared with predecessor state-of-the-art models based on Recurrent Neural Networks (RNN). 
Transformers use a complicated architecture with encoder and decoder layers to understand sequences of words and provide an output \cite{DBLP:journals/corr/VaswaniSPUJGKP17}.
While the encoder layer processes input text data, extracting hierarchical representations through mechanisms like self-attention, the decoder layers generate output sequences based on the input received from the encoder. 
Transformer-based LLMs include GPT models like Generative Pre-trained Transformer 3 (GPT-3), GPT-4 and GPT-4o developed by OpenAI. 
GPT-4 and GPT-4o are multimodal models that accept text and image and produce text \cite{chatgpt2024}. 
Other transformers include \textit{Bidirectional Encoder Representations from Transformers} (BERT) and its smaller and lighter version of \textit{DistilBERT}, designed for applications with limited computational resources. 
BERT and DistilBERT are also designed to understand context in language processing and are suitable for NLP tasks like text classification, answering questions, and named entity recognition. 
The difference between models like BERT and GPT is the way their architecture is designed and the intended learning objectives. 
%BERT uses only the encoder of the transformer architecture and processes text bidirectionally by considering both preceding and following words. 
%On the other hand, GPT models use only a decoder of the transformer architecture and process text in an unidirectional manner by processing text based only on preceding words. 

LLMs can assist in analysing large amounts of text data and identify patterns automatically, which can be helpful when dealing with fraud and other crime-related data.
LLMs have been successfully applied in various areas of human communications, including chatbots in customer support systems, by generating human-like text, content generation, and performing language translation. 
\textit{Generative AI} (GenNAI or GAI) refers to AI techniques that create new text, audio, images and video that closely resemble human-generated content. 
On the other hand, criminals can misuse these resources to generate content for fraudulent activities, such as fake websites, targeted phishing emails, and scam advertisements, to deceive potential victims.

\subsection{The Use of AI in Fraud Detection}
Although some literature reviews explore the application of AI for fraud and crime, to the best of our knowledge, no reviews currently aim to understand the state-of-the-art in detecting online fraud in general. The literature reviews we found, discuss the detection of specific online fraud or scams, such as credit card fraud~\cite{cherif2023credit} and phishing SMS~\cite{barrera2023literature}, among others.

In more detail, our preliminary analysis of literature reviews finds that specific AI models work best towards detecting specific types of fraud (e.g., phishing URLs, smishing, etc.), as researchers perform literature reviews to analyse specific offences and not analyse the general task of fraud overall.
In addition, a single/universal model does not perform well at classifying various types of fraud. 
Hence, researchers must constantly develop and update their trained models to detect specific fraud types.
In this work, we aim to understand whether there are \emph{universal} AI methodologies that attempt to detect online fraud, in general, focusing on textual data.

\begin{table*}[t]
    \centering
    \begin{tabular}{l|l|l}
    \hline
        \textbf{Search String} & \textbf{Library} & \textbf{Notes}\\
        \hline
        
         ("Online Fraud*" OR "scam*")& ACM & All text \\ 
         AND (("machine learning") & ProQuest & All text, Journals, Conferences\\ 
         OR "NLP" OR ("natural language processing")  & Web of Science & Topic \\
         OR "classifier" OR ("Large Language Models") OR "LLM" &IEEE Xplore&Abstract, Journals, Conferences\\ 
         OR ("Generative Artificial Intelligence") & arXiv&Abstract, Computer Science, Jan 2023-Mar 2024\\ 
         OR "GenAI" OR "GAI") & Google Scholar & All text, Review articles, 2023-2024\\ 
        \hline
    \end{tabular}
        \caption{Search query for the literature selection in various academic libraries. Notes depict the advanced search filters applied to each library.}
        \label{tbl:search_string}
\end{table*}

\descr{Research Questions:}
\begin{mycompactitem}
    \item \textbf{RQ1}: What is the state-of-the-art of AI techniques used to detect online fraud?
    \item \textbf{RQ2}: What are the data sources researchers use to analyse online fraud? 
    \item \textbf{RQ3}: How do researchers evaluate their AI models?
    \item \textbf{RQ4}: What are the most popular fraud activities that researchers studied?
\end{mycompactitem}

\noindent Although a wide number of studies have explored the application of AI for fraud detection and other types of cybercrime, we are unaware of any systematic literature reviews that have examined the application of AI models using text data. 
This SLR focuses on AI-based models that study textual data to detect and gain insights about online fraud. 
Thus, this study identifies NLP models used to detect online fraud.

\section{Systematic Review Methodology}\label{sec:systematic-review}
Systematic reviews differ from traditional literature reviews as they aim to identify all relevant studies that address a set of research questions using a structured methodology that can be replicated~\cite{nightingale2009guide}.

\subsection{Methods}
We use the following methodology to conduct the SLR and address the selection process to identify relevant publications and avoid biases.

\subsubsection{Protocol}
We followed the Preferred Reporting Items for Systematic Reviews and Meta-Analysis extensions for Scoping Reviews (PRISMA), as proposed by Moher et al.~\cite{moher2010preferred}. In a nutshell, this provides a comprehensive framework for conducting and reporting systematic reviews and meta-analyses. The process includes a checklist and flow diagram to ensure transparency, reproducibility, and rigour in summarizing research evidence, to improve the quality of reviews in various fields and to standardise how a literature review should be reported.

\begin{table*}[!ht]
    \centering
    \begin{tabular}{p{2cm}|p{7.2cm}|p{5cm}}\hline
        \textbf{Label} & \textbf{Description} & \textbf{Example}\\\hline
        Title & The title of the manuscript & Detecting Phishing URLs Using NLP \\\hline 
        Author & Author’s full name plus the abbreviation et al. if applicable & Smith, John or Smith et al. \\\hline 
        Year & The year the work got published (YYYY) & 2020 \\\hline 
        Fraud Type & The type of scam the authors try to detect, analyse, or discuss in their manuscript & Phishing URLs \\\hline 
        Data Type & The type of data the authors use for their analysis & URLs \\\hline 
        Data Quantity & The amount of data used for the analysis & 100 phishing URLs, 100 legitimate URLs \\\hline 
        Models Used & All models the authors experimented with & RF, LDA, W2V \\\hline 
        Best Model & The model with the best accuracy & Random Forest \\\hline 
        Model Stats & All performance metrics & A=0.95, P=0.81, R=0.9, AUC=0.89 \\\toprule
    \end{tabular}
        \caption{Data items and characteristics extracted from the literature.}
        \label{tbl:data_extracted}
\end{table*}

\subsubsection{Eligibility}\label{sec:eligibility}
This review focuses on studies that use AI-based models, specifically NLP models, including Machine Learning (ML) and Deep Learning (DL) techniques using text data. 
For example, studies that employ AI to detect fraudulent bank accounts, fraudulent credit card transactions, or fraudulent networks of users online were out of scope.
For a study to be considered for inclusion in the SLR, we used the following eligibility criteria: 
\begin{mycompactitem}
\item \textbf{Peer-Reviewed Studies}: We focused on peer-reviewed studies published in English between January 2019 and March 2024. Our search was restricted to academic records found in journals and conference proceedings. We excluded theses, legal documents, patents, and citations.
\item \textbf{Grey Literature}: To capture the latest AI-based models, we also included grey literature, specifically pre-prints from ArXiv published between January 2023 and March 2024 that have not yet been incorporated into conference proceedings or academic journals. 
We also conducted searches on Google Scholar between January 2023 and March 2024. 
\item \textbf{Search Strategy}: Table~\ref{tbl:search_string} shows the search string used to query related literature in ACM Library, ProQuest, Web of Science, IEEE Xplore, arXiv, and Google Scholar. This was finalised after trying various searches in these libraries. Due to the different functionalities of each library, we had to adjust our search accordingly: for ACM library, we queried our search string across the entire text and adjusted the time range; for ProQuest, we queried our search string across the entire text, adjusted the time range, included only papers from conferences and journals, included only full text and peer-reviewed results, and filtered the subjects to exclude medical terms; for Web of Science we queried our search string on the topic (title, abstract, and keywords) of the paper and adjusted the time range; for IEEE Xplore, we queried our search string on the abstract of the papers, adjusted the time range, and filtered results for papers published in conference proceedings and journals; for arXiv, we queried our search string on the abstract of papers, published in computer science between January 2023 and March 2024; and finally, for Google Scholar, we queried our search string and filtered our results on review articles published between 2023 and 2024.
The adjustments mentioned above were implemented to better capture literature related to the scope of our study, and a consensus was reached after various iterations and discussions between all of the authors.
\item \textbf{Scope and Focus}: Studies must address fraud performed online and use AI-based methodologies for analysing online fraudulent activities. The focus was on studies using \emph{textual data}, whether from scammers, victims, or victim reports, to understand, detect, or analyse online fraud activities. Our goal was to understand the state-of-the-art models designed to prevent and detect scams before the victim gets defrauded. Studies that analysed money transactions, credit card transactions, and cryptocurrency transactions were \textit{excluded} from this review, as they do not use text data.
\item \textbf{AI-Methodology}: Papers had to include a methodology or similar section where the authors discuss their AI implementation and fine-tuning along with the accuracy of their model. Finally, we considered studies published after 2019.
\item \textbf{Publication Time Frame}: Papers published between January 2019 and March 2024 were included in this review. This period was selected to manage the overwhelmingly large volume of online fraud papers and align with our available resources. Also, we believe that studies conducted before 2019 are less likely to reflect recent advancements in AI methods. Given the rapid evolution of AI-based models, our time frame ensures the inclusion of the most up-to-date and relevant research.
\end{mycompactitem}

\subsubsection{Data extraction}
Next, the authors agreed on the data to be extracted from the included studies. 
Only one of the three reviewers carried out the task of extracting data. 
This was deemed sufficient since the reviewer's role involved extracting the required details from the papers, and a second reviewer did not have to check accuracy. 
The only aspect of the data extraction that the reviewer had to conceptualize was the specific \emph{Fraud Type} analysed by the study under question. For example, if a study analysed Phishing URLs, it was labelled as \emph{Phishing URLs}.

Not all papers explicitly specified the type of fraud analysed. 
Due to the diversity of scams, there is no agreed way of labelling fraud types. 
As a result, we used an umbrella term to categorise them. 
For example, online scam campaigns made by bot users on various social media platforms to advertise fraudulent phishing URLs include \emph{fake users} and \emph{phishing URLs} analysis; hence, we agreed to label papers of broad online scam campaigns as \emph{social media scams}.

We did not classify a paper with more than one fraud type to ease the representation of our findings.
Instead, the reviewer recorded the paper's primary goal when identifying the fraud analysed. 
For example, if a paper used phishing emails to extract phishing URLs towards detecting phishing URLs, then that paper would be labelled as \emph{Phishing URLs}, as that was the study's main goal. 
The final version of the data extracted from each record is depicted in Table~\ref{tbl:data_extracted}, along with relevant examples.
A thematic analysis was conducted on the extracted data of the studies included for qualitative analysis, and we present our findings in Section~\ref{sec:summary-of-findings}.

%{\color{darkred}
%\subsubsection{Limitations and Challenges}
%Despite our best efforts to ensure the robustness and inclusivity of our systematic review methodology, we encountered several challenges and limitations.

%One major challenge was adjusting our search filters on the different academic libraries to return homogeneous results while remaining relevant to the scope of our study. 
%In addition, the different academic libraries and databases required tailored adjustments to their filters, and they do not provide the same filter options.
%For example, we had to search our string on the entire text (\emph{all}) of the papers of the ProQuest library as it does not provide a choice to search the string on the \emph{topic} of the paper and when queried on the \emph{abstract} alone, it would fail to retrieve any results.
%In addition, Google Scholar often failed to return consistent results when applying the exact search string using its advanced search functionality. 
%Since Google Scholar's advanced search functionality proved unreliable for complex queries like ours, we had to search the entire database. 
%This sometimes resulted in an unmanageable number of results, complicating the selection process.

%Similarly, for arXiv, the limitations of its advanced search capabilities led us to implement a custom Python script to query the database and retrieve relevant publications using our complex search string. 
%While this solution was effective, it required additional effort and technical expertise.}

\section{Results}\label{sec:slr_results}

\subsection{Study selection and characteristics}
\begin{figure}[t]
\centering
\includegraphics[width=1\columnwidth]{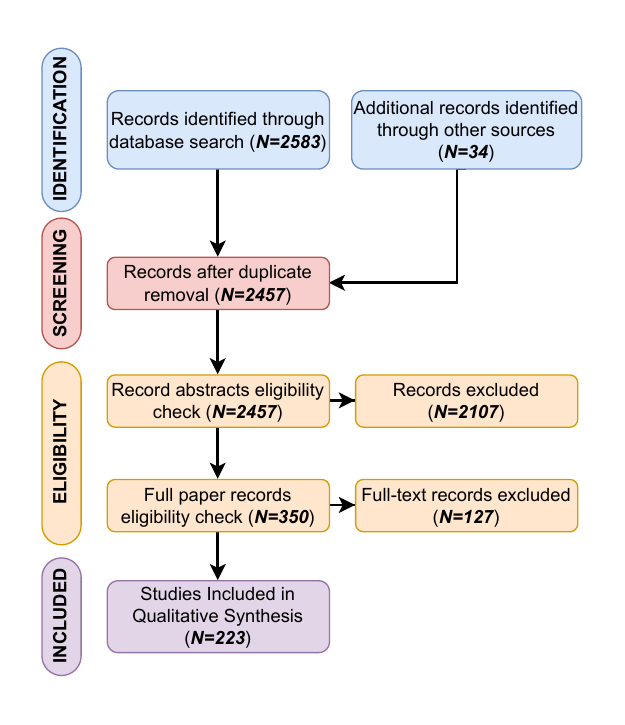}
\caption{PRISMA Chart}
\label{fig:prisma_chart}
\end{figure}

The PRISMA-ScR flow diagram in Figure~\ref{fig:prisma_chart} summarises the study selection process. 
We identified $2,617$ studies for eligibility screening. 
The ACM Digital Library returned $389$ documents, IEEE Xplore returned $712$ documents, Web of Science returned $253$ documents, ProQuest returned $783$ documents, Google Scholar returned $399$ documents, and ArXiv returned $47$ documents. 
Experts in the area recommended an additional $34$ papers. 
After removing duplicates, $2,457$ papers remained for further review.

At this stage, $10\%$ of these papers were selected ($N=242$) for Inter-Rater Reliability to calculate the multi-annotator agreement between the three annotators of this review. 
The Fleiss Kappa score between all three annotators was $0.83$, indicating almost perfect agreement. 
The Cohen Kappa Agreement was also calculated between each pair of annotators. 
The agreement between annotators AP and NT was $0.65$ (substantial agreement), between AP and EM was $0.66$, and between NT and EM was $0.52$ (moderate agreement).\footnote{AP stands for author Antonis Papasavva, NT for author Nilufer Tuptuk, and EM for author Enrico Mariconti.}
The three annotators compared their annotation process and reviewed this SLR's eligibility criteria and goals. 
Then, the lead annotator performed the rest of the annotations of the papers included in this review.

Reviewing the abstract of those papers resulted in $2,107$ papers being excluded from the study as they did not fit the eligibility criteria discussed in Section~\ref{sec:eligibility}. 
This resulted in $350$ full-papers passing to eligibility screening, out of which $127$ did not fit the eligibility criteria and were excluded.
Overall, this process resulted in $223$ full-text papers being included for qualitative analysis.

\begin{figure*}[ht!]
\centering
\includegraphics[width=0.8\textwidth]{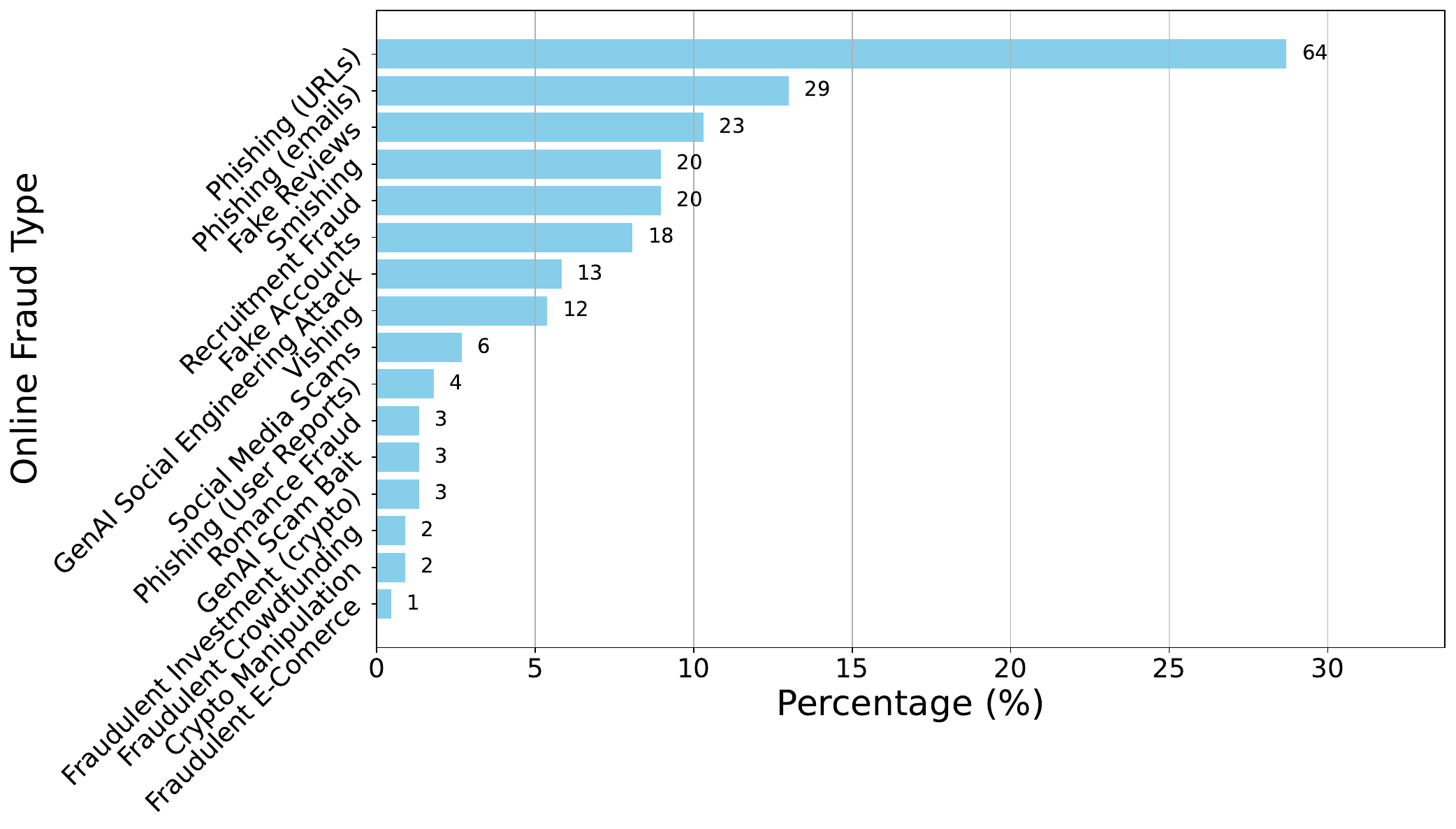}
\caption{Percentage of scam types analyzed in the studies included for qualitative analysis}
\label{fig:percentage_scam_types}
\end{figure*}

\subsection{Types of Online Fraud Identified in the Literature}
Figure~\ref{fig:percentage_scam_types} shows the types of fraud analysed in the full-text papers included in the qualitative analysis. 
The reviewer of these studies manually coded each paper with the \emph{scam type} that the study focuses on, based on its title, abstract, and methodology. 
The majority of studies focus on \textit{phishing} detection, with about a third ($29\%$) of the studies analysing phishing URLs online ($N=64$).
More specifically, these works tackle the problem of automatically detecting whether a URL is likely fraudulent.
Many papers were related to detecting phishing emails ($N=29$), followed by studies on SMS phishing detection ($N=20$).
Other studies on phishing include phone call transcripts towards understanding and detecting voice call phishing ($N=12$), and a few studies attempt to understand phishing methods via victim reports ($N=4$).

Moving on to other types of fraud, we found many studies that detect fake reviews on various platforms like the Google Play Store, Apple App Store, Yelp, and TripAdvisor ($N=23$).
Another widely studied scam was \textit{recruitment fraud} ($N=20$). 
We also found several studies that employed AI techniques to detect 
\textit{fake accounts} on Facebook, Instagram, and Twitter ($N=18$).

Similarly, $3$ studies focus on \textit{romance fraud} via analysing profiles on social media and discussions with victims.
We also identified $3$ studies that analysed and detected fraudulent \textit{cryptocurrency investment} scams and $2$ studies that attempted to detect the likelihood of cryptocurrency manipulation.
Finally, we identified $2$ studies that analysed fraudulent crowdfunding online and $1$ that studied fraudulent e-commerce websites.

\descr{Studies on other emerging types of fraud.}
A few studies have used  Generative AI (GenAI) to analyse and better understand existing and emerging fraud techniques, especially ones where GenAI or LLMs are misused towards social engineering attacks.
We identified $3$ studies that employed GenAI models to automatically interact with scammers to waste their resources and time while gathering information on the methods they used to defraud users, thereby disrupting their operations. 
This approach is defined as \emph{scam baiting}: the process of using generative AI models to deceive and engage with scammers. 
Notably, this is a \emph{countermeasure} against online fraud (``GenAI Scam Bait'' in the Figure~\ref{fig:percentage_scam_types}). 

Our search also returned many studies that discuss the exploitation of \emph{GenAI towards social engineering attacks} ($N=13$), where scammers use these advanced models to create legitimate-looking targeted emails or SMSs to earn the trust of potential victims. 
Although GenAI models offer numerous benefits, these studies show the significant risks they pose when used for malicious purposes, particularly in social engineering. 
GenAI can generate coherent, contextually relevant, and grammatically correct emails that mimic the style and tone of professional communication. 
This increases the likelihood of victims perceiving fraudulent emails as legitimate and trusting the message~\cite{schmitt2023digital}. 

Regarding other scams, we found $6$ studies that detect \textit{social media scams}. These scams included various fraudulent activities, like fake user accounts and online groups, advertisements of fraudulent apps, or phishing websites that aim to trick users into exposing their personal information or paying money.

The above three groups of studies were not identified in the literature discussed in Section~\ref{sec:online-fraud}; hence, we grouped and discussed them briefly here.

\section{Summary of Findings}\label{sec:summary-of-findings}
This section summarises our findings, categorized per fraud activity analysed within the papers included in our SLR for qualitative analysis.

\subsection{Data Sources}
First, we report the most popular data sources used, and the datasets analysed.

\descr{Data Used for Phishing URL Detection.}
We start by understanding the chosen data sources for analysing and detecting phishing URLs; the most popular scam-type category we have detected in our SLR.
We analysed the data sources and detection methodologies of the identified $63$ papers that focused on this issue. 
Table~\ref{tbl:data_phishing_URL} summarizes these.

Researchers used various websites that offer information on URLs for the analysis of malicious and legitimate domains. 
This information may be web page rankings (how trusted the webpage is), phishing reports, and historical data.
By far, the most popular data source used was PhishTank\footnote{\url{https://phishtank.org/}}, a website that allows users to report webpages that might be malevolent or suspicious, with $25$ studies using it as already labelled malicious websites~\cite{li2022mui,vo2023shark, al2020convolutional, aslam2023phish, jishnu2023enhanced, jaber2022improving, rafsanjani2023qsecr, alswailem2019detecting, mandadi2022phishing, jha2023intelligent, marimuthu2022intelligent, adebowale2023intelligent, shaiba2022hunger, rao2020two, salloum2021phishing, orunsolu2022predictive, rao2019detection, barraclough2021intelligent, almseidin2019phishing, chiew2019new, li2019stacking, sahingoz2019machine, somesha2020efficient, tharani2020understanding, do2020malicious, pradeepa2022lightweight}.

Another website that offers a list of phishing URLs is OpenPhish\footnote{\url{https://openphish.com/}} and it was used in $3$ studies~\cite{vo2023shark, almseidin2019phishing, chiew2019new}.
Two studies used URLhaus\footnote{\url{https://urlhaus.abuse.ch/}}, a project for sharing malicious URLs, for the collection and analysis of phishing URLs~\cite{rafsanjani2023qsecr, do2020malicious}.
We also find one study that used SpamHaus\footnote{\url{https://www.spamhaus.org/}} for the collection of IP and domain reputation~\cite{fernandez2022early}, and one that used URLscan\footnote{\url{https://urlscan.io/}}~\cite{chen2021ai}.
Interestingly, a study ~\cite{chen2021ai} also collected user-reported domains from ScammerInfo\footnote{\url{https://scammer.info/}}, a forum where users post and discuss various scams.
Lastly, the webpage WhoIs\footnote{\url{https://who.is/}}, a webpage that offers historical data on webpages, was used for feature collection in two studies~\cite{shalke2022social,adebowale2023intelligent}.

The most used data source for the collection of legitimate webpages was ``Alexa,'' (a global ranking system that estimated a website's popularity that shut down in May 2022) with $10$ studies using it to collect legitimate annotated webpages~\cite{li2022mui,navyah2022ensemble, saha2023phishing, liang2019using, shaiba2022hunger, orunsolu2022predictive, rao2019detection, somesha2020efficient, tharani2020understanding, do2020malicious}.
Google's search engine was used for one study~\cite{chen2022development}, while another used the Majestic Million site\footnote{\url{https://majestic.com/reports/majestic-million}} for legitimate webpage collection~\cite{jishnu2023enhanced}, a site similar to Alexa.

Many studies used existing publicly available datasets for their analysis.
More specifically, $11$ studies~\cite{gu2022ensemble,jha2023machine,jha2023machine, mehndiratta2023malicious, jain2023support, saha2020phishing, kumar2023hybrid, ashwitha2023perception, dr2023malicious, pradeepa2022lightweight, zamir2020phishing} used publicly available datasets published on Kaggle (a repository for researchers to publish data).
Similarly, $5$ studies~\cite{navyah2022ensemble, jaber2022improving, kalabarige2023boosting, mohammed2022accuracy,priya2020gravitational} used the UCI Phishing Dataset.\footnote{\url{https://archive.ics.uci.edu/dataset/327/phishing+websites}}
Other studies used publicly available datasets from other sources~\cite{ariyadasa2022phishrepo,villanueva2022application,vecile2022malicious,kumar2020phishing,zin2023machine,pradeepa2022lightweight,pathak2023classification}.

The most recent studies (published in 2023) that attempted to detect phishing URLs automatically collected data from alternative sources like social networks and user-reported phishing URLs~\cite{bitaab2023beyond,saha2023phishing,saha2023phishing,bitaab2023beyond,nakano2023canary,janet2022real}, while others used datasets from telecom and Security organizations~\cite{mehndiratta2023malicious,yu2024efficient,liang2019using}.
Alas, we failed to detect the data source used by $9$ studies, as the authors did not report how or from where they acquired the dataset used in their study~\cite{alkawaz2022identification,el2021malweb,kundra2023identification,chen2020intelligent,ou2021no,raja2021lexical,yadollahi2019adaptive,nagy2023phishing,geyik2021detection}.

\descr{Data Used for Phishing Email Detection.}
We now discuss the data sources used in the $29$ works that tackle phishing email detection.
The data extracted from the literature and presented in this section are shown in Table~\ref{tbl:data_phishing_emails}.

The overwhelming majority of papers used datasets made available in previous work~\cite{al2022digital,bhatti2021email,janez2023review,stojnic2021phishing,saka2022context,jena2023malicious,jonker2021using,janez2021trustworthiness}, or used datasets published on Kaggle~\cite{janez2023review,janez2021trustworthiness,chataut2024can,markova2019classification,al2020email,islam2021spam,ramprasath2023identification,singh2022spam,emmanuel2023information,livara2022empirical}, or datasets published at UCI ML repository~\cite{salihovic2019role,ismail2022efficient,kushwaha2023analysis,saini2023machine,mittal2023blockage}.

Two studies~\cite{genc2021understanding,jiang2024detecting} used emails received in the author’s personal or professional email spam folder.
Another study that included datasets from alternative sources was by Mehdi et al.~\cite{mehdi2023adversarial}. 
They used various techniques to develop their dataset, including GPT2 generated synthetic phishing emails made available in previous research~\cite{radford2019language}, along with TextAttack\footnote{\url{https://github.com/QData/TextAttack}}, a Python framework for adversarial attacks, data augmentation, and model training in NLP, Textfooler\footnote{\url{https://github.com/jind11/TextFooler}}, a Model for Natural Language Attack on Text Classification, and Probability Weighted Word Saliency (PWWS)~\cite{ren2019generating}, a technique for generating adversarial text. 

Another alternative data source for phishing email detection was used by Janez et al.~\cite{janez2021trustworthiness} who used data from SPAM Archive\footnote{\url{http://untroubled.org/spam/}}, a website that publishes spam email repositories at the end of every month and is constantly updated.	
Gallo et al.~\cite{gallo2019identifying} analysed user-reported emails.
Lastly, the data source used in $3$ studies was not clearly stated within the manuscript~\cite{mughaid2022intelligent,venugopal2022detection,rahmad2020performance}.

\descr{Data Used for Phishing SMS Detection.}
Regarding Phishing SMS (\emph{smishing}), we included and analysed $20$ papers in this SLR.
For the detailed data, refer to Table~\ref{tbl:data_smishing}.

Similarly to previous analyses, the overwhelming majority of works opted for using already publicly available datasets to analyse and train a model.
More specifically, a variety of subsets from a publicly available dataset on Kaggle\footnote{\url{https://www.kaggle.com/datasets/uciml/sms-spam-collection-dataset}} was used by $14$ studies~\cite{vinothkumar2022detection,agrawal2023effective,jain2019feature,mishra2023dsmishsms,abid2022spam,kohilan2023machine,siagian2023improving,gandhi2023sms,al2023multi,zhang2023bert,dharani2023spam,addanki2023safeguarding,ulfath2021hybrid,jain2022sms}.
Another study~\cite{zhang2020lies} used the Kaggle dataset but incorporated Fake Base Station data and made it available to researchers.\footnote{\url{https://github.com/Cypher-Z/FBS_SMS_Dataset }}
Similarly, this work~\cite{vinothkumar2022detection} used a subset of the Kaggle dataset in combination with emails and YouTube comments for spam content detection, while Lai et al.~\cite{lai2022semorph} used data provided by users.\footnote{\url{https://www.datafountain.cn/competitions/508}}
Tang et al.~\cite{tang2022clues} collected tweets where users reported smishing for their analysis.
Two other works used data from the Korean Internet and Security Agency~\cite{seo2024device} and 360 Mobile Safe~\cite{liu2021detecting}.
Lastly, Timko et al.~\cite{timko2023commercial} proposed a platform where users can freely post Phishing SMS for researchers to use.\footnote{\url{https://smishtank.com/}}

\descr{Data Used for Phishing Phone Call Detection.}
We identified $12$ studies that used phone call transcripts to understand voice call-enabled phishing (\emph{vishing}).

Derakhshan et al.~\cite{derakhshan2021detecting} used the CallHome dataset, which includes 120 unscripted 30-minute telephone conversations between native speakers of English.\footnote{\url{https://catalog.ldc.upenn.edu/LDC97S42}}
Another study~\cite{djire2023evaluating} used AI-generated deepfake voice recordings (Tacotron 2\footnote{\url{https://pytorch.org/hub/nvidia_deeplearningexamples_tacotron2/}}, Deepvoice 3\footnote{\url{https://r9y9.github.io/deepvoice3_pytorch/}}, and FastSpeech 2~\cite{ren2020fastspeech}).
For authentic voice recordings, they used the synplaflex dataset~\cite{sini2018synpaflex}, a corpus of audiobooks in French.

Some studies used telecommunication operator datasets, like~\cite{liang2023telecom} using fraudulent caller IDs	and phone transcripts~\cite{huang2022learning} from telecommunication operators in China, Hu and Yuan~\cite{hu2023urf4cct} used data from the Public Security Bureau in Zhejiang Province, China, and~\cite{kim2021voice} used data from the Korean Financial Supervisory Service.
Kale et al.~\cite{kale2021classification} developed their dataset via questionnaires and victim testimonies.
Other authors collected data from various social networks, including YouTube transcripts~\cite{rahman2022phone}, Facebook, online blogs and forums, public datasets, as well as some that were developed based on studies of scammers' activities and behaviours~\cite{hong2023scam}.
Others opted for using previously analysed and publicly available data~\cite{gowri2021detection,malhotra2023detection}, while the data Zhong et al.~\cite{zhong2020encoding} used was unclear.										

\descr{Data Used for Phishing (User Reports) Detection.}
Four studies used user reports to understand phishing activities. First, one study~\cite{liu2021fine} constructed a fraud complaint dataset from the Internet finance service in China.
Similarly, another~\cite{zhou2022keyword} used court documents from Chinese online judgement records, while a third~\cite{palad2019document} used incident record forms from victims and interviews in the Philippines.
In the final study, the authors~\cite{lwin2023supporting} launched and introduced a website operated by the National Crime Prevention Council (NCPC) in Singapore, where users could report and receive information on the latest phishing activities.~\footnote{\url{https://www.scamalert.sg/}}

\descr{Data Used for Fake Review Detection.}
For this type of scam, $23$ studies were included in our analysis, and the data extracted from them is depicted in Table~\ref{tbl:data_fake_reviews}

The overwhelming majority of papers used a previously published YELP dataset\footnote{\url{http://odds.cs.stonybrook.edu/yelpzip-dataset/}}~\cite{javed2021fake,pengqi2023unmasking,harris2022combining,tufail2022effect,balakrishna2022identifying,ashraf2023fake,wang2021modeling,singh2023deep}, or the OTT publicly available dataset on Kaggle\footnote{\url{https://www.kaggle.com/discussions/general/281540}}~\cite{singh2023deep,balakrishna2022identifying}.

Other studies used application reviews from the Google Play Store or Apple's App Store~\cite{yugeshwaran2022rank,tushev2022domain,obie2022violation}.
Other studies used previously available Amazon product reviews\footnote{\url{https://snap.stanford.edu/data/web-Amazon.html}}, or collected Amazon reviews~\cite{rangar2022machine,iqbal2023efficient,furia2020tool,gupta2020detecting,thilagavathy2023fake,chandana2021analyzing,akshara2023small,deekshan2022detection}.

Some studies collected reviews of Amazon Hotel and Holiday packages ~\cite{rangari2022empirical,silpa2023detection}, or TripAdvisor review data~\cite{tufail2022effect}
And, lastly, one study~\cite{bevendorff2024product} used YouTube transcripts to interpret false review exaggeration.	
The data source used by Ganesh et al.~\cite{ganesh2023implementation} was unclear.

\descr{Data Used for Recruitment Fraud Detection.}
We found $19$ studies that focused on the detection of fraudulent job postings (see Table~\ref{tbl:data_recruitment}).

The overwhelming majority ($N=16$) of papers, used the same publicly available dataset from Kaggle,\footnote{\url{https://www.kaggle.com/datasets/amruthjithrajvr/recruitment-scam}}, which holds about 18K job postings of which 800 are fraudulent.
Notably, this dataset includes data from 2012 to 2014~\cite{nessa2022recruitment,prathaban2022verification,pandey2022effective,ranparia2020fake,habiba2021comparative,reddy2023web,santhiya2023fake,lal2019orfdetector,nasser2021online,amaar2022detection,sofy2023intelligent,vo2021dealing,nanath2023investigation,ullah2023smart,bhatia2022detection,li2021exploratory}.

The other three studies developed custom crawlers to collect data from various job posting sites in the UK (SEEK, Glassdoor, Indeed, and Gumtree)~\cite{mahbub2022online}, in Bangladesh (job.com.bd, bdjobstoday, deshijob)~\cite{tabassum2021detecting}, and in China (China-Boss, Zhipin, Liepin, and 51job)~\cite{zhang2023orfpprediction}.

\descr{Data Used for Fake Account Detection.}
Some studies attempted to tackle the automated detection of fake profiles online, and Table~\ref{tbl:data_fake_users} shows the data extracted from them.

We found that many authors collected user profile data from online social networks including Twitter~\cite{raj2020multi,gangan2023detection,yue2019madafe,singh2023safe,ali2019detect,shukla2022tweezbot,rovito2022evolutionary}, Instagram~\cite{das2022effecient,fathima2023ann,anklesaria2021survey}, Facebook~\cite{albayati2019identifying,venkatesan2019graph,shreya2022identification}, YouTube~\cite{na2023evolving}, and Sina Weibo~\cite{zhang2021social}.
A different approach used in one study~\cite{haq2023spammy} was to collect real names from various web pages, schools, and other sources to detect fake names online automatically.

Other authors have used previously published and openly accessible datasets that included user data from various social networks~\cite{nikhitha2023fake,bebensee2021leveraging}.

\descr{Data Used for GenAI Social Engineering Attack Detection.}
Under this category, we found many works that investigated how generative AI models can be misused to defraud people.

Some studies~\cite{janjeva2023rapid,schmitt2023digital,ferrara2024genai} develop and discuss an initial taxonomy for which they discuss how scammers can misuse AI-generated content.
At the same time, Carlini et al.~\cite{carlini2021extracting} test various membership inference attacks – which is when someone attempts to figure out whether a specific piece of data was used to train a machine learning model – on OpenAI's GPT2 model and confirm that the model is vulnerable to this kind of attack which poses risks to privacy.	
Similarly, Kumar et al.~\cite{kumar2023augmenting} discuss the significant implications for cybersecurity, privacy, and ethical considerations that should be considered when developing and using these models.

Apropos misuse cases of these models, Ayoobi et al.~\cite{ayoobi2023looming} discuss how LLMs and GenAI can be used to create fake professional profile bios to trick users into believing that the account is legitimate.
Similarly, DiResta and Goldstein~\cite{diresta2024spammers} show that scammers can use these models to create AI-generated images to be posted on social networks.
Their case study shows that these images tend to receive high volumes of engagement on Facebook as many users do not seem to recognize that the images are synthetic. 	
Other research shows how these models can be \emph{jailbroken}\footnote{Jailbreaking a generative AI model means bypassing its safety rules or restrictions to make it produce responses it’s not supposed to.} to produce code to imitate legitimate webpages (phishing URLs)~\cite{grbic2023social}, malware code, phishing emails, phishing SMSs, SQL injection attacks, and other potentially malicious material~\cite{shibli2024abusegpt,alotaibi2024cyberattacks,alawida2024unveiling}.

Other research suggests that humans may be able to accurately detect phishing AI-generated content~\cite{sharma2023well}, while Roy et al.~\cite{roy2023chatbots} discuss and experiment with countermeasures to prevent malicious prompts (jailbreaking) for GPT and provide insights into how the model can become more robust against this vulnerability.

\descr{Data Used for Social Media Scam Detection.}
Six studies examined various scams and spammers facilitated by Social Networks.
Xu et al.~\cite{xu2022efficiently} used data from WeChat (a Chinese messaging, social media, and mobile payment app) and Konect repository to detect users that use WeChat to defraud people.										
La Morgia et al. (2023)~\cite{la2023sa} and La Morgia et al. (2021)~~\cite{la2021uncovering} used Telegram data to characterize and detect Fake Telegram channels, while Shah et al.~\cite{shah2020illicit} collected data from Telegram and compared it to Twitter data to understand and detect fake users. 
Similarly, Al-Hassan et al.~\cite{al2023dspamonto} collected and analysed Twitter and Institute of Informatics and Telematics data to detect scammers on Twitter.
Finally, Tripathi et al.~\cite{tripathi2022analyzing} collected YouTube data to detect scammers attempting to lure victims using comments posted alongside YouTube videos.	

\descr{Data Used for Romance Fraud Detection.}
In their study, He et al.~\cite{he2021datingsec} attempted to automatically detect malicious accounts on Momo,\footnote{\url{https://www.immomo.com/aboutus.html}} a dating website.
Similarly, Suarez-Tangil et al.~\cite{suarez2019automatically} collected data from \href{datingnmore.com}{datingnmore.com} and \url{scamdigger.com} to develop automated methods to understand fraudulent profiles within dating social networks.
Lastly, Lokanan~\cite{lokanan2023tinder} analysed the sentiment of tweets with the hashtag \#tinderswindler to provide an understanding of users sharing their experiences regarding romance fraud.

\descr{Data Used for GenAI Scam Baiting.}
We identified three studies.~\cite{cambiaso2023scamming,bajaj2023automatic,chen2023active} that used LLMs and Generative AI to automatically engage with scammers online to waste their resources and collect data on various fraud activities.
For these studies, the researchers collected data from their own baiting accounts and emails and said data was not made publicly available.

\descr{Data Used for Fraudulent Investment Detection.}
Studies that attempted to understand fraudulent investment scams employed various datasets and methodologies. 
First, Siu et al.~\cite{siu2022invest} analyse investment scam advertisements found in Bitcointalk.\footnote{\url{https://bitcointalk.org/}}
Li et al. collected YouTube comments to detect bots that advertise fraudulent investment content automatically~\cite{li2023towards}.
Lastly, Kuo and Tsang ~\cite{kuo2024constructing} develop a scam detection model based on emotional fluctuations of user discussions collected from one of Taiwan's most popular instant messaging applications.

\descr{Data Used for Fraudulent Crowdfunding Detection.}
Two studies were identified that analyse fraudulent crowdfunding~\cite{lee2022backers,shafqat2019topic}. 
These collected the descriptions and metadata from hundreds of Kickstarter campaigns.\footnote{\url{https://www.kickstarter.com/}}

\descr{Data Used for Crypto Manipulation Detection.}
Market, and more specifically, cryptocurrency coin manipulation, is when users collectively attempt to alter investor interactions towards manipulating the price of a coin.

Nizzoli et al.~\cite{nizzoli2020charting} discuss this process via data acquired from Twitter, Telegram, and Discord channels.
Similarly, Mirtaheri et al.~\cite{mirtaheri2021identifying} identify and analyse cryptocurrency manipulations from user activity collected from Telegram and Twitter.

\descr{Data Used for Fraudulent E-commerce Detection.}
The only study for this category~\cite{salihovic2019role} analysed the terms and conditions of websites that sell (a variety of) products to inform understanding and the detection of obscured financial obligations in online agreements.

\subsection{Methodologies employed}
We now discuss the most popular AI and NLP methodologies employed to study each type of online fraud.

\descr{Methods Applied for Phishing URL Detection.}
The studies included in our SLR attempted to automatically detect phishing URLs using a variety of NLP and AI methodologies. 
These included classic supervised machine learning algorithms such as Naive Bayes (NB), Random Forest (RF), Decision Tree (DT), Support Vector Machines (SVM), XGBoost (Extreme Gradient Boosting), KNN (k-Nearest Neighbors), as well as Artificial Neural Networks (ANN) and more advanced deep learning approaches such as Long Short-Term Memory (LSTM) and Convolutional Neural Network (CNN). 
LSTMs are Recurrent Neural Networks (RNN) designed to capture long-dependencies in sequential data, making them suitable for handling and predicting text sequences. 
On the other hand, CNNs aim to identify key features in the text by capturing local patterns. In the research reviewed, these models were developed to complete the binary classification task of determining whether a URL was fraudulent or not. 

NLP techniques related to text mining have been used to extract features from URLs, which are then used as features to train an AI-based classification model. 
For example, such features include the counts of the characters, special characters, and n-grams of the URLs.
Also, the authors collected other URL features, such as whether the URL had a secure scheme or not (e.g. https), domain (e.g. amazon.co.uk) and top-level domain (e.g. /kitchen), and sub-directories (e.g. /appliances).
All the above and more features were used to fit and train a malicious URL detection model.
Some studies used hybrid or a combination of methodologies, including more advanced techniques.  
For example, Li et al.~\cite{li2022mui} used a Bidirectional Long Short-Term Memory (Bi-LSTM) recurrent neural network, that could process sequences of text in both forward and backward directions, along with a Visual Geometry Group (VGG) which is a type of CNN.
Vo Quang et al.~\cite{vo2023shark} used a Convolutional Neural Network (CNN), along with features extracted using Word2Vec (W2V), a Gated Recurrent Unit (GRU), which is a more simplified type of RNN than LSTM, and a Bi-LSTM. 
Nakano et al.~\cite{nakano2023canary} used BERT with RF; Bitaab et al.~\cite{bitaab2023beyond} developed a hybrid system that uses RF, SVM, FNN, and XGBoost; Alswailem et al.~\cite{alswailem2019detecting} used Linear Regression (LR) and DT; and, Villanueva et al.~\cite{villanueva2022application} used LSTM and GRU.

Other stand-alone AI methodologies, like LightGBM,\footnote{Light Gradient Boosting Machine - an ensemble learning technique designed for handling large datasets with large features} RF, NB, and ANN, also seem to work well on detecting phishing URLs, but the approach with the best performance seems to be RF.

\descr{Methods Applied for Phishing Email Detection.}
The methodologies used for phishing email recognition focus more on NLP analysis. 
The majority of studies used various NLP methodologies for feature extraction, including, but not limited to topic modeling (LDA, BERT, BERTLARGE), text representation (TF-IDF, BOW, Clustering, W2V), and sentiment analysis (VADER, WordNet).

Studies that also aimed to automate the detection of phishing emails employed LLM analysis, RF, NB, SVM, CatBoost, LSTM, RNN, and many more.

\descr{Methods Applied for Phishing SMS Detection.}
Like phishing email detection, phishing SMS detection relies on state-of-the-art NLP methodologies, including LLMs, LDA, BERT and W2V. 
The existing literature also used AI methodologies like LR, SVM, CNN, GNN, LSTM, NB, and KNN for automated detection. 

\descr{Methods Applied for Phishing Phone Call Detection.}
Studies on vishing detection used various AI methods for automated detection.
Most used transcript text data for their analysis, except for Djir{\'e} et al.~\cite{djire2023evaluating} who analysed deepfake voice analysis. 
In that study, the authors found that RNNs performed best.

Overall, various NLP and AI techniques were used on text data, including but not limited to SVM, NB, LSTM, CNN, RF, BERT, W2V, LR, and KNN. 

\descr{Methods Applied for Phishing (User Reports) Detection.}
We found that the use of BERT, Sequential Minimal Optimization (SMO), J48 (an implementation of decision tree), NB, RF, XGBoost, Doc2Vec (D2v - an extension of Word2Vec), Jaccard, NER (Named Entity Recognition), and TF-IDF was applied to analyse user reports of various phishing activities.

\descr{Methods Applied for Fake Review Detection.}
The most popular NLP technique for fake review detection rely heavily on sentiment detection techniques, including VADER and WordNet.
Similar to previous fraud analyses, the AI methods applied included various neural network models, such as CNN.

\descr{Methods Applied for Recruitment Fraud Detection.}
The techniques employed for the automated detection of Fraudulent Job Postings included stand-alone machine learning algorithms used for classification tasks, including LR, SVM, KNN, RF, XBoost, and ANN, and deep learning models, including Bi-LSTM. 

\descr{Methods Applied for Fake Account Detection.}
The studies included in our SLR leveraged various NLP and AI-based techniques to detect fake accounts on social media platforms such as Twitter, Facebook, Instagram, and YouTube. 
Classic supervised learning approaches, including NB, RF, DT, SVM, LR, KNN, and ANN, were widely adopted. 
Ensemble learning methods such as AdaBoost and stacking models were also explored, along with advanced methods like Gradient Boosting techniques (e.g., CatBoost). 
Notably, RF was often found to deliver the best performance in several studies, e.g.,~\cite{anklesaria2021survey, nikhitha2023fake, das2022effecient}.

Deep learning approaches were also employed, particularly when tackling larger datasets. 
For example, Na et al.~\cite{na2023evolving} used RoBERTa to detect fake accounts involved in scam campaigns on YouTube, while Alhosseini et al.~\cite{ali2019detect} leveraged Graph Convolutional Neural Networks (GCNN) for spam bot detection on Twitter. 
Studies such as~\cite{venkatesan2019graph} adopted unsupervised learning techniques like HDBSCAN for anomaly detection in social networks. 

Across the studies reviewed, researchers have extracted diverse features, including user profile characteristics (e.g., number of followers, account age), content-based features (e.g., hashtags, posts), domain-based features, and behavioural patterns. 
%For instance, \citet{raj2020multi} utilised hashtag-, content-, user-, and domain-based features to classify fake Twitter accounts, finding that RF achieved the highest accuracy. 
%Similarly, \citet{shukla2022tweezbot} proposed TweezBot, an AI-driven bot detection algorithm for Twitter, using various feature sets and classifiers like SVM, NB, DT, and RF, reporting an accuracy of 98\%.

%Hybrid and ensemble methods were also prominent in the reviewed studies. 
%For instance, one study~\cite{bebensee2021leveraging} combined RF with node features (NF) and graph features (GF) to detect bots in Twitter datasets.
%Other novel approaches included evolutionary computation for bot detection~\cite{rovito2022evolutionary}, and semi-supervised learning methods like SSML-CatBoost~\cite{zhang2021social}.

Deep learning models such as Bi-LSTM, GRU, and CNN were less frequently applied but showed promise. 
Fathima et al.~\cite{fathima2023ann} developed an ANN-based system to categorise fake profiles on Instagram. 
Ensemble learning approaches, such as combining ANN, SVM, and RF~\cite{shreya2022identification}, were also utilised to enhance detection performance.

The performance of these methods varied depending on the dataset and feature selection, but overall, RF emerged as the most consistent and accurate classifier for fake account detection across multiple platforms and studies.

\descr{Methods Applied for GenAI Social Engineering Attack Detection.}
GPT-3.5 and GPT-4 were the most commonly used models, particularly in generating phishing emails, malicious websites, and smishing campaigns. 
Studies by Roy et al.~\cite{roy2023chatbots} and Shibli et al.~\cite{shibli2024abusegpt} demonstrated how these models could be exploited to craft highly convincing phishing content, leveraging sophisticated language capabilities. 
Ayoobi et al.~\cite{ayoobi2023looming}, utilised models like BERT, RoBERTa, and Flair to detect fake LinkedIn profiles generated by ChatGPT. 
Defensive mechanisms were also explored; for instance, Roy et al.~\cite{roy2023chatbots} proposed BERT-based countermeasures to mitigate malicious prompt exploitation. 
Despite the promising results in detecting and preventing misuse, other research ~\cite{alotaibi2024cyberattacks} ~\cite{alawida2024unveiling} has highlighted vulnerabilities in GPT-3.5, particularly its susceptibility to jailbreak attempts, which enable the generation of harmful content such as SQL injections, malware, and phishing scams.
Overall, LLMs demonstrated advanced capabilities for deception, and their susceptibility to misuse necessitates robust detection and prevention strategies.
Studies under this category did not report performance metrics as they tested the limits of LLMs and GenAI models using qualitative approaches.

\descr{Methods Applied for Social Media Scam Detection.}
Xu et al.~\cite{xu2022efficiently} proposed the BREAD framework, which uses bidirectional k-hop reachability query processing over dynamic graphs to extract fraud-related features. 
La Morgia et al. (2023)~\cite{la2023sa} studied fake Telegram channels employing a Multilayer Perceptron (MLP) model. 
Shah et al.~\cite{shah2020illicit} applied techniques like W2V, D2V, P2V, and TF-IDF to detect illicit activity. 
Tripathi et al.~\cite{tripathi2022analyzing} examined monetised scam videos on YouTube using RF and W2V. 
Al-Hassan et al.~\cite{al2023dspamonto} developed DSpamOnto, an ontology-based model for social spammers on Twitter, and benchmarked it against classifiers such as NB, SVM, and RF. 
Finally, La Morgia et al. (2021)~\cite{la2021uncovering} used LDA for topic modelling. 

Due to the diversity in the types of fraud analysed across social media platforms and the distinct datasets and methodologies employed, identifying a single best-performing model for this category is not applicable.

\descr{Methods Applied for Romance Fraud Detection.} 
Studies of Romance Fraud detection used various NLP methodologies, including sentiment detection, which uses BOW and textBlob, and statistical methods like TF-IDF, for feature extraction.  
When testing different models, researchers have found that Random Forest (RF) performed best for the detection of this offence~\cite{lokanan2023tinder}.
He et al.~\cite{he2021datingsec} found that LSTM is most effective at detecting malicious accounts in dating applications, while another study~\cite{suarez2019automatically} showed that Ensemble Machine Learning (EML) also works well.

\descr{Methods Applied for GenAI Scam Baiting.}
One of the Scam Baiting studies~\cite{cambiaso2023scamming} used OpenAI's ChatGPT to reply to scammer emails. 
Similarly, Bajaj and Edwards~\cite{bajaj2023automatic} experimented with OpenAI's ChatGPT and DistillBERT to categorize scam emails they received and provided a qualitative analysis of how well the two models performed. 
Chen et al.~\cite{chen2023active} set up an email server as a ``honeypot'' from which they sent emails to scammers (to encourage those scammers to send phishing emails to them) identified in data from the Scambaiter mail server, Enron Email Dataset, and ScamLetters.Info.
They then employed their own semi-unsupervised DistillBERT model to engage with scammers automatically and used their model filtering to categorise and analyse the emails received.

\descr{Methods Applied for Fraudulent Investment Detection.}
Three of the studies related to Fraudulent Investment used data from different sources (forums, messaging apps, and YouTube).
Two of the studies~\cite{kuo2024constructing,siu2022invest} reviewed under this SLR used models to detect emotional fluctuations in discussions between victims and found that DT was the best-performing machine learning model for this task. 
Siu et al.~\cite{siu2022invest} also concluded that XGBoost performed well in terms of detecting malicious advertisements for fraudulent investment websites.

\descr{Methods Applied for Fraudulent Crowdfunding Detection.}
One of the papers on fraudulent crowdfunding detection applied NLP methodologies, including Named Entity Recognition and other NLP features detected in the descriptions of Kickstarter campaigns, and built an LR model that performed well~\cite{lee2022backers}.
The other study~\cite{shafqat2019topic} developed an LSTM-LDA topic detection model that analyses the crowdfunding campaign and people's comments with the aim of estimating whether a campaign was a scam. 

\descr{Methods Applied for Crypto Manipulation Detection.}
One study on cryptocurrency market manipulation used pre-existing methods for detecting fake users, along with CorEx Topic Analysis~\cite{nizzoli2020charting}.
The other study found that SVM with SGD and TF-IDF worked best for detecting discussions that aimed to manipulate the market~\cite{mirtaheri2021identifying}.

\descr{Methods Applied for Fraudulent E-Commerce Detection.}
Finally, the only study that we identified that used text data to inform understanding of Fraudulent E-commerce activities, used OpenAI's GPT-4 model to automatically detect obscure financial obligations in the terms and conditions of the websites sampled~\cite{salihovic2019role}.

\subsection{Key Findings}
We now summarise our findings in relation to the research questions listed in Section~\ref{sec:online-fraud}.
To remind the reader, these questions were: to detect the state-of-the-art AI techniques used to detect online fraud (RQ1); the data sources used (RQ2); how researchers evaluate their AI models (RQ3), and what the most studied fraud activities were (RQ4). The answers to RQs 1-3 are organised by fraud type, while the listing of these fraud crime types addresses RQ4.

\descr{Takeaways: Phishing URLs.}
While established datasets have played a major role in developing phishing URL detection models, there is a clear need to incorporate more dynamic and current data sources. 
Leveraging user-reported phishing URLs from social networks and data from telecom and security organizations would offer a more effective approach to combating phishing attacks. 
These sources provide real-time, diverse, and relevant data that enhance the robustness and accuracy of detection models, keeping pace with the evolving nature of phishing threats.
By combining the strengths of both traditional and modern data sources, researchers can develop more comprehensive and adaptive phishing detection systems, better protecting users from phishing URLs.

Regarding the methodologies used, we find that the existing literature used state-of-the-art methodologies to analyse and detect phishing URLs.
Of these, RF seems to be the stand-alone model that works best, while other hybrid methodologies also report promising performance.
Regarding performance reporting, authors often fail to adequately report all of the performance metrics of their model.
Although the Accuracy of the model is reported in all but seven studies, other metrics like Precision, Recall, F1, and AUC are omitted in more than half of the studies we analysed for this type of online fraud.

\descr{Takeaways: Phishing Emails.}
While publicly available datasets have laid the groundwork for phishing email detection research, the rapidly evolving nature of phishing attacks requires the use of more dynamic and up-to-date data sources. 
Leveraging user-reported emails, real-time spam collections, and advanced synthetic data generation techniques could significantly enhance the robustness and accuracy of phishing detection models. 
By combining traditional datasets with innovative data sources, researchers could develop more comprehensive and adaptive phishing detection systems that are better equipped to detect phishing activities via email.

All of the studies that developed automated approaches to phishing email detection reported very good performance, with RF, BERT, LSTM, RNN, and SVM being the most popular.
Similar to the Phishing URLs above, accuracy was the metric reported most often.
Only two studies reported AUC, and Precision, Recall, and F1 were rarely reported.

\descr{Takeaways: Phishing SMS.}
This type of scam was also found to rely on existing datasets.
Alas, the rapidly evolving nature of smishing requires a more dynamic and diversified approach to data collection. 
Researchers could develop more effective and resilient smishing detection models by integrating publicly available datasets with real-time, user-reported data and specialized security sources. 
This approach would ensure that models remain relevant and capable of addressing new and sophisticated smishing threats as they arise.

Contrary to phishing email detection, we find that in the case of phishing SMS detection, which tends to involve much less text, SVM and various applications of Neural Networks performed best.
Again, the Precision, Recall, F1-score, and AUC metrics were underreported, with Accuracy being the metric most studies report.

\descr{Takeaways: Phishing Phone Calls.}
Regarding the data sources used for automated vishing detection, most studies used text data obtained by transcribing voice recordings.
Other research also used caller ID information from various telecommunication operators.
Some researchers collected data from user reports, and only one attempted to detect deepfake signals in voice recordings.
The best-performing technique used for automated vishing detection was SVM.
Although Accuracy was also the most reported performance metric for this category, many studies failed to provide any metrics.

\descr{Takeaways: Phishing (User Reports).}
Reviewing the four studies that focused on phishing via user reports, we found that the researchers used data from various sources, including court judgements, public data from forums, and user reports from Financial Institutions.
The applied methodologies varied and included Named Entity Recognition, various NLP and statistical techniques (D2V, Jaccard, TF-IDF), and ML techniques (SMO, J48, NB, RF, XGBoost). 
Only one study~\cite{liu2021fine} provided performance metrics, and this was for their BERT model that performed best.

\descr{Takeaways: Fake Reviews.}
Although many studies used previously available datasets to establish and test their detection models, we noticed a clear trend where more recent studies tended to collect data from platforms like Amazon, Google Play, the Apple App Store, and YouTube.
This is very encouraging as the data used for these detection models need to be constantly updated.
Hybrid models, including LR, SVM, CNN, and LSTM, seemed to perform best in the detection of fake reviews.
Again, the Precision, Recall, F1, and AUC metrics were underreported, with Accuracy being the metric most studies report.

\descr{Takeaways: Recruitment Fraud.}
While the Kaggle dataset has been pivotal in advancing research on fraudulent job postings, the rapidly evolving nature of job scams necessitates the use of more current and diverse data sources. 
Custom data collection methods, which tap into active job posting sites, represent a critical step forward in enhancing the effectiveness of detection models. 
Researchers can develop more comprehensive systems to effectively combat fraudulent job postings by leveraging a mix of established and new data sources.
The models that performed best for this fraudulent activity varied.
However, we find that Bi-LSTM, KNN, RF, DNN, and LightGBM performed well.
All but one study reported the accuracy performance metric of their best-performing model, while the other four metrics (Precision, Recall, F1, and AUC) remain underreported.

\descr{Takeaways: Fake Accounts.}
Most studies utilised user profile data from popular social networks such as Twitter, Instagram, Facebook, YouTube, and Sina Weibo. 
Several works relied on openly accessible datasets published from previous studies, and others collect user data through APIs, web scraping, or manual curation. 
Advanced techniques such as RoBERTa, CatBoost, and Graph Convolutional Neural Networks (GCNN) have been employed alongside traditional classifiers like Random Forest, Support Vector Machines, and Neural Networks, with Random Forest being one of the most popular and frequently high-performing models.

Performance across studies is generally strong, with reported accuracy often exceeding 90\%, though other metrics like Precision, Recall, and F1 are underreported in some works. 
However, the rapidly changing strategies used by fake account creators highlight the need for more dynamic datasets and adaptive models to tackle this challenge effectively.

\descr{Takeaways: GenAI Social Engineering Attacks.}
Many studies discuss how GenAI models might affect cybersecurity and privacy, the ethical issues they pose, and how they could be misused to create fraudulent content automatically.
However, only two studies identified in our review used data collected from real use cases. 
These were Ayoobi et al.~\cite{ayoobi2023looming}, who discussed fake profile AI-generated content on professional social networks, and Diresta et al.~\cite{diresta2024spammers}, who examined deepfakes posted on Facebook.
No performance metrics were reported in these studies as they were not applicable.
Authors studying this offence were not building models but rather evaluating or experimenting with existing tools, including OpenAI's ChatGPT.

\descr{Takeaways: Social Media Scams.}
The detection of social media scams presents unique challenges due to the diversity of platforms, fraud types, and datasets. 
Most studies leverage platform-specific data sources such as Telegram~\cite{la2023sa,la2021uncovering,shah2020illicit}, WeChat~\cite{xu2022efficiently}, Twitter~\cite{al2023dspamonto,shah2020illicit}, and YouTube~\cite{tripathi2022analyzing} to build detection models. 

The methodologies employed in these studies vary widely, encompassing advanced NLP techniques (W2V, P2V, and LDA)~\cite{shah2020illicit,la2021uncovering}, as well as machine learning classifiers (RF, NB, and SVM)~\cite{tripathi2022analyzing,al2023dspamonto}. 

While these studies report promising results, this category involved studies on various kinds of social media scams. 
Hence, the lack of standardisation across models and datasets limits the generalisability of these findings. 
Notably, these studies also often underreported evaluation metrics like AUC and F1.

\descr{Takeaways: Romance Fraud.}
Romance fraud detection has primarily focused on analysing user-generated content on dating platforms and social media. 
Studies utilised diverse datasets, including user profiles from dating platforms~\cite{he2021datingsec,suarez2019automatically} and tweets tagged with \#tinderswindler~\cite{lokanan2023tinder}. 
Methods employed feature extraction techniques like BOW, TF-IDF, and sentiment analysis with textBlob~\cite{lokanan2023tinder}. 
Among machine learning models, RF consistently performed well~\cite{lokanan2023tinder}, while LSTM achieved the best performance in detecting malicious accounts in dating apps~\cite{he2021datingsec}, and EML yielded high accuracy in identifying fraudulent profiles~\cite{suarez2019automatically}. 
Due to the limited number of studies under this category, we cannot make conclusions regarding the models' overall performance reporting and generalisability.

\descr{Takeaways: GenAI Scam Baiting.}
The use of GenAI for scam baiting has shown promising potential in wasting scammers’ resources and collecting data for fraud analysis. 
Studies employed LLMs such as ChatGPT~\cite{cambiaso2023scamming,bajaj2023automatic} and semi-unsupervised DistillBERT~\cite{chen2023active,bajaj2023automatic} to engage with scammers and categorise phishing emails. 
These studies highlighted the potential of GenAI tools in automating scam baiting, but future work should focus on creating publicly available datasets and refining engagement strategies to improve scalability and efficacy.

\descr{Takeaways: Fraudulent Investment Detection.}
The detection of fraudulent investment scams has been explored using a variety of data sources, including Bitcointalk~\cite{siu2022invest}, YouTube~\cite{li2023towards}, and instant messaging apps~\cite{kuo2024constructing}. 
However, the small sample of just three studies limits the ability to conclusively identify the best-performing model for this category. 
The studies reported varied performance metrics, with Siu et al.~\cite{siu2022invest} highlighting XGBoost as the top performer for detecting fraudulent investment advertisements on Bitcointalk. 
Kuo and Tsang~\cite{kuo2024constructing} found DT to be the best model for identifying emotional fluctuations in scam discussions on a popular Taiwanese messaging app, and Li et al.~\cite{li2023towards} focused on arbitrage bot scams and utilised NNs for their analysis, without reporting performance metrics. 
As the three studies under this category analysed different aspects of a fraudulent investment, we cannot draw any conclusions regarding the best-performing model.

\descr{Takeaways: Fraudulent Crowdfunding.}
Fraudulent crowdfunding detection has been explored using a small set of two studies, both focusing on Kickstarter campaigns~\cite{lee2022backers,shafqat2019topic}. 
LR, in combination with many NLP features, was employed to identify fraudulent campaigns, achieving an accuracy of 87.3\%. 
A combination of LSTM and LDA based on user comments was also reported, without reporting performance metrics. 
Given the limited scope of these studies, further research is required to assess the robustness of these models in detecting fraudulent crowdfunding activities. 
Both studies highlight the effectiveness of NLP-based approaches for feature extraction and classification in this domain.

\descr{Takeaways: Crypto Manipulation.}
The detection of cryptocurrency market manipulation was investigated in two studies, which used data collected from Twitter, Telegram, and Discord~\cite{nizzoli2020charting,mirtaheri2021identifying}. 
Methodologies used included network analysis, topic analysis, and SVM.
The limited number of studies made it difficult to assess the robustness of the models, and further research is needed to identify the best-performing AI technique for this type of fraud.

\descr{Takeaways: Fraudulent E-Commerce.}
The only study identified for fraudulent e-commerce detection~\cite{salihovic2019role} used OpenAI's GPT-4 model to analyze the terms and conditions of e-commerce websites and detect obscure financial obligations, such as shipping, subscription, and refund fees. 
However, the study did not provide performance metrics or compare different models. 
Due to the limited nature of this single study, it is impossible to draw conclusions about the approach's effectiveness or the model's general applicability in detecting fraudulent e-commerce websites.

\section{Discussion}\label{sec:discussion}
We now discuss our findings, discussing recognised limitations and shortcomings identified in the reporting of AI models related to the performance and data sources used. We also provide recommendations for researchers developing detection models for online fraud. 

\subsection{Data Sources}
Overall, we analysed the data sources used and the detection methodologies employed in $223$ papers that aimed to address a range of online fraud problems. 
Although our findings reveal a preference for well-established datasets, especially in the automated detection of various phishing and fake reviews detection, more recent studies (published after 2023) seem to shift towards more dynamic and recent sources.

We found that the overwhelming majority of studies concerned with the detection of phishing relied on well-known and extensively studied datasets from websites like PhishTank, OpenPhish, and SpamHaus. 
In contrast, others used datasets made available from previous studies or publicly available repositories like Kaggle, the University of California Irvine (UCI) Machine Learning Repository, and GitHub. 
This was also the case for studies that focused on phishing email detection, fake review detection, and fraudulent recruitment detection. 
While these established datasets provide a valuable foundation for research, they come with limitations. 
 
Online fraud is dynamic, with new scam techniques continually evolving, or building over older scam methods.
Studies show that LLM-empowered bots, or scammers could be deployed to generate and automate sophisticated and targeted fraudulent and phishing content online, either in the form of email, a professional profile, or deceptive terms and conditions for fake e-commerce websites~\cite{janjeva2023rapid,schmitt2023digital,ferrara2024genai}. 
Hence, relying on outdated datasets may limit the effectiveness of detection models when applied to current or evolving threats. 
The historical datasets used do not capture the latest trends and variations in the different online fraud techniques and activities we see today (and will see tomorrow).

Recent studies have started to leverage more dynamic and real-time data sources to address these limitations. 
Regarding automated phishing detection, recent studies used user-reported phishing URLs from social networks and data from telecom and security organizations. For instance, studies published in 2023 utilized data from sources like Twitter, Facebook, and specialized security organisations~\cite{bitaab2023beyond,saha2023phishing,saha2023phishing,bitaab2023beyond,nakano2023canary,janet2022real,mehndiratta2023malicious,yu2024efficient,liang2019using}.

At the same time, recent research on automated phishing email detection has utilised user-reported emails, providing a real-time perspective on phishing threats~\cite{gallo2019identifying}. 
For example, Genc and Jiang\cite{genc2021understanding,jiang2024detecting} used emails from their personal or professional spam folders, capturing a more realistic and up-to-date snapshot of phishing attacks.
Mehdi et al.\cite{mehdi2023adversarial} took an innovative approach by incorporating various techniques to develop their dataset; GPT-2 generated synthetic phishing emails and tools like TextAttack, TextFooler and PWWS. 
This approach provides a diverse dataset and ensures the model is robust against sophisticated phishing techniques.
In their study, Janez et al.~\cite{janez2021trustworthiness} used data from the SPAM Archive\footnote{\url{http://untroubled.org/spam/}}, which is a continuously updated repository of spam emails.

Similarly, several studies concerned with automated phishing SMS detection have extended the datasets used by combining pre-existing publicly available datasets with additional sources to improve the robustness and generalizability of their models.
This has included the use of additional data from Fake Base Stations~\cite{zhang2020lies}, emails, YouTube comments~\cite{vinothkumar2022detection}, user-reported data for research~\cite{lai2022semorph} or content posted on Twitter~\cite{tang2022clues}.
In particular, Timko et al.~\cite{timko2023commercial} proposed a platform, SmishTank, where users can post phishing SMS messages, creating an ongoing and up-to-date repository for researchers.
We also observed studies using data from specialised security agencies~\cite{seo2024device} and mobile security services~\cite{liu2021detecting}, which offer a more targeted collection of smishing examples.

On Fake Review detection, one study~\cite{bevendorff2024product} used YouTube transcripts to interpret false review exaggeration, showcasing an innovative approach to identifying fraudulent content in multimedia contexts.
Others have adopted similar techniques and developed their own data collection methodologies to collect reviews from sources like App stores, e-commerce websites, and location and travel research platforms. 

An overwhelming number of studies focused on fraudulent recruitment detection using the same dataset from Kaggle.\footnote{\url{https://www.kaggle.com/datasets/amruthjithrajvr/recruitment-scam}} 
Only three studies employed custom crawlers to gather data from various job posting websites in different countries, providing a more diverse and up-to-date perspective. 
Mahbub et al.~\cite{mahbub2022online} collected data from job posting sites in the UK, Tabassum et al.~\cite{tabassum2021detecting} gathered job postings from Bangladeshi sites, and Zhang et al.~\cite{zhang2023orfpprediction} sourced data from Chinese job sites.
The use of up-to-date data collection from these sources offers some advantages. 
For one, these studies can capture the most recent and relevant data by scraping data from active job posting sites, reflecting current fraudulent practices.
Second, collecting data from multiple sources across different regions provides a richer and more varied dataset, which can enhance the robustness and generalizability of detection models.
Finally, custom datasets often include a wider variety of job postings, including niche or less common types of employment scams, which can be critical for developing more comprehensive detection systems.

Overall, combining publicly available datasets with \emph{recent} data from other sources, such as social media, user reports, and specialized agencies, may significantly enhance the robustness and relevance of detection models. 
This approach could ensure that models are exposed to a wider variety of fraud tactics and can adapt to new threats more effectively.

There are various advantages of using such dynamic data sources: 
\begin{mycompactitem}
    \item Real-time Updates: Social networks and security organizations provide continuously updated data. This ensures that the detection models are trained on the most recent phishing URLs, making them more robust against new and emerging threats.
    \item Diverse Data: User-reported data from social networks and institutions often include a wide variety of phishing techniques and strategies. This diversity enhances the model's ability to generalize and detect a broader range of phishing attacks.
    \item Early Detection: These sources can help in the early detection of new phishing campaigns. Social networks, in particular, can act as early warning systems where new phishing URLs are often first reported.
    \item Enhanced Relevance: Data from telecom and security organizations are often more relevant to current threats and can include targeted phishing attacks that are not present in older datasets.
\end{mycompactitem}

However, despite the advancements in data collection methods, there are still gaps. 
For instance, the data sources used in studies for several types of online fraud were not clearly stated within the manuscripts. 
This lack of transparency can hinder reproducibility and the ability to compare results across different studies.

\subsection{Methodologies}
The methodologies employed across various studies of phishing and fraudulent activities involved a wide range of AI that incorporate NLP, machine learning and deep learning techniques. Most of these techniques involved the extraction of features using NLP techniques and then applying supervised machine learning (i.e. models that use labelled data) and deep learning algorithms to build binary classifiers. 

For phishing URLs, Machine Learning and Deep learning algorithms such as CNN, ANN, KNN, LSTM, NB, RF, DT, SVM, and XGBoost were commonly used, often with URL feature extraction through NLP methods like character counts and n-grams. 
We also found various studies that applied hybrid approaches combining multiple techniques, demonstrating strong detection capabilities. 
Similarly, phishing email detection heavily relied on NLP for feature extraction (e.g., LDA, BERT) followed by AI models like RF, NB, and SVM for classification. 
Similar techniques were applied for phishing SMS detection. 

For vishing (phishing phone calls), transcript analysis was primarily conducted using NLP and AI models, with some studies examining deepfake voice detection. 
The analysis of user reports about phishing activities utilized models like BERT and RF, while studies of fake reviews often involved sentiment analysis using methods like VADER. 

Fraudulent job posting detection has involved both the use of machine learning and deep learning models such as LR and Bi-LSTM, while  
romance fraud detection has used sentiment detection methods in combination with machine learning models (e.g. RF) and deep learning models (e.g. LSTM). 
Classic machine learning models like DT, XGBoost, and SVM were employed for fraudulent investment and crypto manipulation. 
Studies on fraudulent e-commerce and crowdfunding leveraged advanced NLP and machine learning techniques, including GPT-4 and LR, respectively.

Although the research and detection methodologies applied in the reviewed literature performed well, they are not without limitations.
Overall, popular machine learning algorithms like RF and SVM often rely heavily on the quality of extracted features, which can be labour-intensive to generate and may miss out on subtle indicators when dealing with large data sets. 

In addition, complex models based on deep learning techniques like Bi-LSTM with VGG or hybrid approaches can be computationally expensive and difficult to implement in real-time systems due to the large amount of data processing resources that they require.

Notably, models developed that work well for one kind of fraud might not be generalised well to other fraud activities. For example, unlike emails, SMS messages are typically short, providing limited data for accurate feature extraction and classification.
In addition, natural language processing techniques used may struggle to capture those features necessary to understand the context (semantics) or syntax of phishing content, leading to potential false positives/negatives (i.e., misclassifications). 

Models trained on specific languages or datasets may not perform well on emails in other languages or different styles.
Many of the challenges that may arise regarding the trained AI models are often the result of poor data quality. 
More specifically, effective feature extraction is critical but can be difficult due to the varied nature of how natural language is used.
At the same time, the textual content used in online fraud activities - such as fraudulent emails, SMSs, or job postings -  keeps evolving, making it challenging for static models to remain effective over time. 

Although collecting data from various websites, forums, social networks, and telecommunication operators for online fraud detection is invaluable, at the same time, different platforms hold data inconsistently or have unique features and user behaviours, complicating model generalization.
Also, identifying fraudulent activities in real-time is challenging due to the dynamic nature of these scams and the lack of real-time occurrences in the various data sources.

In short, while these methodologies offer powerful tools for detecting phishing and fraudulent activities, there are challenges related to feature extraction, model complexity, generalizability, and data quality.
Finally, one of the major issues highlighted in many of these studies was that the models use supervised machine learning models, which require labelled data. 
Creating labelled data is often challenging and time-consuming. 
This is one reason why so many researchers use existing labelled data, but, as discussed, such data will become less useful as fraud evolves over time. 

\subsection{Recommendations}

\descr{Datasets.}
The reliance on older, established datasets for training AI-based models is a double-edged sword. 
While they offer a solid foundation for model development and facilitate the comparative analysis of different models, their static nature may limit their effectiveness in detecting evolving or emerging fraud trends, hence limiting their effectiveness in detecting new types of fraud in the real world. 
Therefore, a strong case exists for incorporating more dynamic and diverse data sources. 
Recent studies that use custom crawlers to gather data from various online platforms that focus on a range of fraud types exemplify best practices in this area. 
These approaches provide real-time, relevant data that can significantly improve the adaptability and accuracy of detection models. 
Going forward, it is recommended that researchers consider combining established datasets with freshly collected data to create more robust and resilient AI models.

\descr{Methodologies Used.}
Overall, most studies reviewed used stand-alone machine learning and deep learning models to detect online fraud. 
In many cases, NLP techniques used for feature extraction were under-reported or ignored. 
Working on online fraud activities that involve textual data should utilise more sophisticated NLP techniques such as transformer-based models (e.g., BERT, GPT) for deeper semantic understanding and better context handling. 
Although LLMs have limitations, such as generating hallucinated or inconsistent results, they are extremely powerful for context extraction.

Recent research demonstrates the utility of hybrid models, which combine different AI and NLP techniques to leverage their strengths. 
These hybrid models appear to perform very well. 
Most existing studies have used supervised machine learning models that require labelled data to detect fraudulent activities. 
Due to the challenges of obtaining new labelled data, researchers often rely on existing datasets that may not capture the content of new techniques and tactics that scammers employ. 

To address these challenges, further exploration of active learning, semi-supervised learning and anomaly-based models that rely on small amounts of labelled data or no labelled data is needed.  
For example, unsupervised or semi-supervised anomaly detection techniques could be studied to identify outliers and novel fraud patterns that may not be present in the training data.
Finally, we observed that almost none of the models reported had real-time applications. There should be a shift in focus where researchers attempt to optimize models for real-time processing to ensure the timely detection and mitigation of fraudulent activities.

\descr{Model Performance Reporting.}
While assessing the literature reviewed in this SLR, we observed many studies that only reported a subset of model performance metrics, and authors frequently relied on the Accuracy performance metric alone.
This was the case for all of the online fraud types identified and analysed herein.
However, using accuracy on its own, especially when the dataset is unbalanced, can be misleading. In a dataset where one class has more observations than another (for example, having fewer phishing emails compared to not-phishing emails), a model could achieve very high accuracy simply by predicting the majority class (i.e. not-phishing emails) without doing a good job of detecting the phishing emails.

Overall, it is essential that researchers report a more comprehensive range of performance metrics beyond accuracy alone. 
These should include Precision, Recall, the F1-score, AUC score, or ROC curve.  These metrics provide a more complete and nuanced picture of a model's performance, especially when dealing with imbalanced datasets. 
In addition, there is a need to conduct and report detailed error analysis to identify common failure cases and the reasons behind them. This can help understand the limitations of the model and areas for improvement. 
Finally,  models need to be cross-validated to ensure the robustness of the reported performance metrics.  For example, reporting results from multiple folds of data samples can provide a more reliable estimate of model performance.

\descr{Reproducibility.}
Many studies failed to explain key and critical aspects of their model development. 
This included the features engineered and selected, methods used for extracting features, the data used, the size of the dataset, the partition of the data into training-test sets, and the hyper-parameters used for tuning and training the models. 

To this end, we recommend researchers provide access to the code, datasets, and pre-trained models used in their studies through platforms like GitHub, GitLab, or institutional repositories. 
This would help improve the reproducibility of their work.
Researchers should also ensure that the methodology section of their paper is sufficiently detailed to allow others to replicate their model and critically assess it. 
This should include a clear description of pre-processing steps, feature engineering and selection, model hyperparameters,  the training-testing data split and training protocols. Furthermore, the use of standardized frameworks and libraries for model implementation (e.g., TensorFlow, PyTorch) could improve reproducibility. 
Comprehensive documentation and setup instructions will help others understand and reproduce the work more easily.

\descr{Usability.}
Most of the papers reviewed were proof-of-concept studies, so the usability of AI-based models has not been addressed. The effective application and use of AI-based approaches depend on successful usability studies that enable users to develop these models into toolkits and provide user feedback. Usability goals are generally determined by efficiency, effectiveness, engagement, error tolerance and ease of use. 
It is thus also imperative to ensure collaboration between the developers of AI-based tools and practitioners. However, while the field of technology usability assessment in front-line policing is growing~\cite{1894801,ZAHABI2018161}, there is a lack of usability studies considering the use of AI in preventive policing, including its application to cybercrimes like online fraud.

\descr{Bias.}
The majority of studies do not discuss the limitations of their models or data. 
Researchers should clearly identify and report such limitations, including any assumptions made, potential biases in the data, and methodologies' limitations. 
The overuse of existing labelled datasets could impact the performance of the models, potentially leading to issues such as over-fitting. 
However, issues like this were not discussed in most of the studies reviewed here.
Researchers should examine the distribution of different classes and any potential sources of bias in the data used for training and testing.
Lastly, there is limited discussion on the generalisability of the models across different datasets, contexts, and evolving fraud patterns. 
Hence, researchers should conduct sensitivity experiments to evaluate the generalisation performance of their models. 

\descr{Limitations.}
As with every study, ours comes with potential limitations.
One that stands out lies in the possibility of missed studies. 
Although we took great care in designing and refining our search string to capture as many relevant publications as possible, the diversity and rapid growth of literature in this field make it likely that some studies were inadvertently omitted. 
Furthermore, grey literature sources such as pre-prints may lack consistent metadata, which could have further limited our ability to identify all relevant studies.

Finally, the inherent subjectivity in labelling fraud types and selecting the primary focus of each paper poses a challenge. 
While efforts were made to standardise the categorisation process via discussions between all authors, bias or misclassification could have introduced slight inconsistencies into the analysis.
In more detail, the challenges associated with classifying online fraud have been extensively discussed in various reports and academic papers.
Rabitti et al.~\cite{rabitti2024taxonomy} and Eling et al.~\cite{eling2021cyber} explain that the ﬁeld of cyber risks is rapidly expanding, and many taxonomy-based systems have been proposed. 
At the same time, the partial taxonomies produced by various bodies have resulted in various reports, often not in harmony with one another, which can lead to many frauds being misclassified or the introduction of grey areas and uncertainty. 
This lack of uniformity in the academic literature renders the identification of online fraud a challenge that is still to be addressed, calling for further research.
Similarly, Cohen et al.~\cite{cohen2019investigation} highlight the lack of consistency across said taxonomies and models, emphasizing the need to understand this risk better. 

Moving on to literature and taxonomies produced outside academia, a review from the UK's National Fraud Authority~\cite{fraudtypologiesuk} explains that online frauds are diverse and can be differentiated further, highlighting the evolving nature of online fraud. That review demonstrates how diverse online fraud can be, taking many forms. 
It compares how different taxonomies and literature reviews use different umbrella terms to classify specific scams and online fraud.
Lastly, a white paper by the UK Police Foundation~\cite{skidmore24} highlights that ``Fraud is daunting in terms of its scale and variety''.
The report discusses in detail the various methods adopted by fraudsters, the exploited criminal opportunities, and the experiences of the victims. 
Notably, the author explains that the online fraud landscape is continuously evolving as fraud methodologies and fraudsters adapt to new technological, social, and commercial opportunities before explaining that online fraud is not defined concretely and is often underreported by victims.
We acknowledge that while undertaking this systematic literature review, we also faced and confirmed the above challenges regarding the evolving nature and inherent issue of classifying online fraud, which likely affected how we classified the studies selected for qualitative analysis and, hence, the presentation of our analysis and final results.

\section{Conclusion}\label{sec:conclusion}
In this systematic literature review, we have examined a wide range of studies focusing on the detection of various fraudulent activities using AI-based models and Natural Language Processing techniques. Our goal was to examine the current state-of-the-art models and techniques used for development and training, investigate the sources of data used, and assess how these models are evaluated. 
Due to resource limitations, we restricted the SLR data collection to 2019-2024. 
The studies we identified covered a wide range of fraudulent activities, but there was a particular focus on phishing attacks. 
However, there is growing interest in using more advanced, Generative AI content to create deceptive content and tools that can be used for scam-baiting. 

Significant attention has been given to building classification models that could be used to detect fraudulent activities. 
In particular, hybrid models that combine advanced NLP techniques with deep learning, including LLMs, have been developed. However, there remains considerable room for improvement. The key AI-model development areas that require attention include performance reporting, reproducibility, and transparency. 
Providing detailed performance reporting will help us to compare and evaluate different models. 
Improving reproducibility is important and requires researchers to provide sufficient details about what they did and how they did it. 
Increasing transparency means providing clear information on how the AI-based models work and make decisions. 
This will help fraud practitioners to interpret and understand the models, and mitigate any biases in AI-based models. 

Furthermore, most existing models rely heavily on labelled data and supervised machine-learning techniques. 
Future studies should give some attention to the application of unsupervised and semi-supervised machine learning for detecting fraud. 
Similarly, the data sources used for training these models are unsuitable for capturing the dynamic nature of fraud. 
Future studies should, therefore, investigate building AI-based models that can capture emerging fraudulent patterns and their usability in fraud prevention. 

Addressing these gaps is crucial for creating more robust, reliable, fair, and ethical AI-based systems to detect fraud. 
By considering the recommended practices, the research community can help to better understand, prevent, detect, and mitigate fraudulent activities and reduce victimisation, ultimately contributing to a safer and more secure digital environment.

\section*{Appendix}
Here we list tables, based on specific scam types, discussed in the body of this manuscript.
For the complete data extracted from the academic works included in this review, please refer to the public repository.\footnote{\url{https://osf.io/nrx7y/?view_only=ca1050d48c4c4a969817c6d5f677cb87}}

\onecolumn
\begin{landscape}
{\footnotesize
\begin{longtable}{|l|p{4cm}|p{2.8cm}|p{3.3cm}|p{2.5cm}|l|l|l|l|l|}
\hline
\multirow{2}{*}{\textbf{\#}} & \multirow{2}{*}{\textbf{URL Source}} & \multirow{2}{*}{\textbf{Collection Method}} & \multirow{2}{*}{\textbf{Models Used}} & \multirow{2}{*}{\textbf{Best Model}} & \multicolumn{5}{|c|}{\textbf{Performance}}\\ 
\cline{6-10}
& & & & & \textbf{P} & \textbf{R} & \textbf{A} & \textbf{F1} & \textbf{AUC} \\ 
\hline
\cite{alswailem2019detecting}(2019) & \url{phishtank.org}* & Custom crawler &RF & RF & & &0.98& & \\ \hline
\cite{rao2019detection}(2019) & \url{phishtank.org} and Alexa* & UNK& J48, RF, SMO, LR, MLP, BN, SVM, AdaBoost & RF & 0.99& &0.99& & \\ \hline
\cite{almseidin2019phishing}(2019) & \url{phishtank.org} and \url{openphish.com}* & Previous work~\cite{chiew2019new}&BNET, NB, J48, LR, RF, MLP & RF & & &0.98& & \\ \hline
\cite{chiew2019new}(2019) & Alexa, \url{phishtank.org}, \url{openphish.com}, and \url{commoncrawl.org}* & UNK&RF, SVM, NB, C4.5, JRip, PART & RF & & &0.94& & \\ \hline
\cite{li2019stacking}(2019) & Alexa and \url{phishtank.org}*&UNK & SVM, KNN, DT, RF, GBDT, XGBoost, LGB, Hybrid (GBoost, XGBoost, and LightGBM) & Hybrid (GBoost, XGBoost, and LightGBM) & & &0.97& & \\ \hline
\cite{yadollahi2019adaptive}(2019) & UNK & UNK & C4.5, AdaBoost, KNN, RF, SMO, NB& Hybrid (XCS/UNK) &0.98& &0.98&0.98&0.99 \\ \hline
\cite{sahingoz2019machine}(2019) & \url{phishtank.org}, Yandex Search API, and GitHub & Open dataset~\cite{github-url-phishing} and custom crawler&DT, AdaBoost, Kstar, KNN, RF, SMO, NB & DT & 0.96& &0.97&0.97& \\ \hline
\cite{liang2019using}(2019) & \href{https://blog.netlab.360.com/}{NetLab360} and Alexa* & UNK&LR, SVM, LSTM & LSTM & 0.98& & 0.98& & \\ \hline
\cite{zamir2020phishing}(2020) & Kaggle&Open dataset~\cite{kaggle-url-data-kumar} & NB, KNN, SVM, RF, Bagging, NN & Hybrid (NN, RF, and Bagging) & 0.95&0.98&0.97&0.96& \\ \hline
\cite{somesha2020efficient}(2020) & Alexa and \url{phishtank.org}*  & Previous work~\cite{rao2019detection}&DNN, LSTM, CNN & LSTM & & &0.99& & \\ \hline
\cite{tharani2020understanding}(2020) & Alexa, \url{phishtank.org}, Mendeley, \url{openphish.com}, and \url{commoncrawl.org}* & Open dataset~\cite{mendeley-url-data} and custom crawler&NB, SVC, KNN & SVC, KNN  &  & & & &\\ \hline
\cite{do2020malicious}(2020) & \url{phishtank.org}, \href{https://urlhaus.abuse.ch/}{URLHaus}, \href{https://majestic.com/reports/majestic-million}{Majestic}, Kaggle & Open datasets~\cite{kaggle-url-data-anthony,majestic-data,urlhaus-data} and custom crawler&SVM, RF & RF &0.98&0.97&0.99& & \\ \hline
\cite{kumar2020phishing}(2020) & Refer to open dataset & GitHub open dataset~\cite{githubURLdataset} & NB & NB & 1&0.95& &0.97& \\ \hline
\cite{saha2020phishing}(2020) & Kaggle* & UNK open dataset & MLP&MLP & & &0.93& & \\ \hline
\cite{al2020convolutional}(2020) & UNK* & Previous works~\cite{mohammad2014predicting,mohammad2014intelligent}& RF, RNN, CNN & CNN & & &0.94& &0.91 \\ \hline
\cite{priya2020gravitational}(2020) & UCI ML Repository & Open dataset~\cite{UCI-ml-phishing-url}& RF, DT, ANN, KNN & RF & & &0.95& &  \\ \hline
\cite{rao2020two}(2020) & \url{phishtank.org} and Google Search* & UNK& SVM, DT, LR, RF, XGBoost, AdaBoost, ET & Hybrid (RF, XGBoost and ET)  & & &0.98& & \\ \hline
\cite{chen2020intelligent}(2020) & \url{phishtank.org}* & UNK & KNN & KNN& & &0.98& & \\ \hline
\cite{raja2021lexical}(2021) & Kaggle and \href{https://www.unb.ca/cic/datasets/url-2016.html}{Canadian Institute of Cybersecurity}* & UNK  & SVC, LR, KNN, NB, RF & KNN & 0.98&0.98&0.98&0.98& \\ \hline
\cite{chen2021ai}(2021) & \url{scammer.info} and \url{urlscan.io}* & Custom crawler & LGBM & LGBM & 1 & 0.96&0.98&0.97&0.98 \\ \hline
\cite{geyik2021detection}(2021) & UNK*&Previous work~\cite{rao2020catchphish}&RF, DT, NB, LR & RF & & &0.83& &  \\ \hline
\cite{el2021malweb}(2021) & UNK*&UNK & LR, DT, NB & LR, DT & 1&1&1&1& \\ \hline
\cite{ou2021no}(2021) & Alexa and \url{cryptoscamdb.org}* & Custom crawler& NB, SVM, KNN, RF & RF & 0.98&0.95&0.97&0.96& \\ \hline
\cite{salloum2021phishing}(2021) & Alexa, \url{phishtank.org}, and Mendeley* & Custom crawler and open dataset~\cite{mendeley-url-phishing} & XBoost, RF, SVM, KNN, ANN, LR, DT, NBB & ANN &0.96&0.97&0.97&0.97& \\ \hline
\cite{barraclough2021intelligent}(2021) & \url{phishtank.org}, \url{relbanks.com}, and \url{millersmiles.co.uk}* & Custom crawler& ANFIS, NB, PART, J48, JRip & PART & 0.98&0.99&0.99&0.99&0.98 \\ \hline
\cite{chen2022development}(2022) & Google Rankings and \url{whoscall.com} & Custom crawler &RF, DNN & RF & 1&0.99&0.99& & \\ \hline
\cite{li2022mui}(2022) & Alexa and \url{phishtank.org}* & Custom crawler & Bi-LSTM, Hybrid (Bi-LSTM and CNN), Hybrid (Bi-LSTM and VGG)  & Hybrid (Bi-LSTM and VGG) & &0.96&0.96&0.96& \\ \hline
\cite{shalke2022social}(2022) & \url{who.is}* & Custom crawler& BPNN, RBFN, SVM, NB, DT, RF, KNN & NB & & &0.96& & \\ \hline
\cite{alkawaz2022identification}(2022) & UNK* & UNK & DT, RF & RF & & &0.8& & \\ \hline
\cite{navyah2022ensemble}(2022) & Alexa, \href{https://archive.ics.uci.edu/}{UCI}, \url{phishtank.org} and Kaggle* & UNK open dataset& KNN, RF, DT, CBoost, LGBM, ABoost, VC & CBoost &  & &0.98&0.98& \\ \hline
\cite{jaber2022improving}(2022) & \url{phishtank.org} and UCI ML Repository & Custom Crawler and open dataset~\cite{UCI-ml-phishing-url} &Hybrid (CNN) & Hybrid (CNN) & & &0.97& & \\ \hline
\cite{vecile2022malicious}(2022) & \href{https://www.unb.ca/cic/datasets/url-2016.html}{Canadian Institute for Cybersecurity} & Open dataset~\cite{cicURLdataset}& LSTM & LSTM & 0.99&0.99&0.99& & \\ \hline
\cite{gu2022ensemble}(2022) & Kaggle* & UNK & RF, KNN, XGBoost & XGBoost & & & 0.96&0.96& \\ \hline
\cite{mandadi2022phishing}(2022) & \url{pishitank.org}* &Custom crawler & RF,DT & RF & & &0.87& & \\ \hline
\cite{villanueva2022application}(2022) & GitHub& Open dataset~\cite{github-url-phishing}&  LR, NB, LSTM, GRU & LSTM or GRU& & &0.95& & \\ \hline
\cite{mohammed2022accuracy}(2022) & UCI ML Repository & Open datasets~\cite{UCI-ml-phishing-url} and UNK& AdaBoost, CART, GBoost, MLP, SVM, RF, NB, SEM & SEM & & & 0.98& & \\ \hline
\cite{marimuthu2022intelligent}(2022) & \url{phishtank.org} and Alexa*&Custom crawler&DT, RF & RF & &  &0.87& & \\ \hline
\cite{fernandez2022early}(2022) & Farsight SIE~\cite{fss}, \url{spamhaus.org}, and \url{surbl.org}* & UNK& J48, RF & RF  & & & & &  \\ \hline
\cite{shaiba2022hunger}(2022) & UNK* & Previous work~\cite{rao2020catchphish}& Hybrid (CNN and LSTM) & (CNN and LSTM) & 0.98&0.99&0.99&0.99&0.99 \\ \hline
\cite{ariyadasa2022phishrepo}(2022) & Mendeley and previous works & Mendeley open dataset~\cite{mendeley-phishrepo-data} and previous works~\cite{lin2021phishpedia,feng2020web2vec} & Hybrid DLM, Stack model, URLNet & Hybrid DLM & & &0.93&0.93& \\ \hline
\cite{orunsolu2022predictive}(2022) & \url{phishtank.org} and Alexa* & UNK& SVM, NB & Hybrid (UNK) & 0.99&0.99& &0.99&0.99 \\ \hline
\cite{pradeepa2022lightweight}(2022) & \href{https://www.unb.ca/cic/datasets/url-2016.html}{Canadian Institute of Cybersecurity}, \url{phishtank.org}, and Kaggle* & UNK&SVM, RF & RF&0.99&0.99& 0.99& & \\ \hline
\cite{janet2022real}(2022) & Twitch*&Twitch API & XGBoost, RF, NB & RF &0.93&0.93& &0.93& \\ \hline
\cite{vo2023shark}(2023) & \url{phishtank.org} and \url{openphish.com} & Custom crawler& CNN & SharkEyes (CNN, W2V, GRU, Bi-LSTM) & 0.94&0.94&0.95&0.94& \\ \hline
\cite{nakano2023canary}(2023) & Tweets, \url{spamhunter.io}, and \url{tweetfeed.live}* &Twitter APi and custom crawler & Hybrid (BERT and RF) & Hybrid (BERT and RF) & 0.96 & &0.95&0.95& \\ \hline
\cite{saha2023phishing}(2023) & Twitter and Meta's \url{crowdtangle.com}* & Twitter API and custom crawler & UNK & Pre-trained model~\cite{li2019stacking} & 0.96&0.97&0.97&0.96& \\ \hline
\cite{jha2023machine}(2023) & Kaggle* & Data no longer available & RF, LR, KNN & RF & 0.97&0.99& &0.97& \\ \hline
\cite{bitaab2023beyond}(2023) & \url{reddit.com/r/Scams/} and \href{https://docs.paloaltonetworks.com/advanced-url-filtering/administration/url-filtering-basics/url-categories}{Paolo Alto Networks}*&Custom crawler & RF, XGBoost, SVM, FFNN & BeyondPhish (RF and XGBoost and SVM and FFNN) & & &0.98& & \\ \hline
\cite{mehndiratta2023malicious}(2023) & Kaggle and \href{https://www.unb.ca/cic/datasets/url-2016.html}{Canadian Institute of Cybersecurity} & Open datasets~\cite{kaggleURLdataset,cicURLdataset} & KNN, LR & KNN & & &0.9& &  \\ \hline
\cite{jain2023support}(2023) & Kaggle* & UNK open dataset& DT, KNN, RF, SVM & SVM & 0.99&0.96&0.98&0.97& \\ \hline
\cite{zin2023machine}(2023) & Mendeley & Open dataset~\cite{mendeley-url-data} & RF, J48, NB, KNN, LR & RF&0.97&0.9&0.94& & \\ \hline
\cite{kumar2023hybrid}(2023) & Kaggle & Open dataset~\cite{kaggle-url-data} & DT, KNN, RF, GBoost & UNK hybrid & & &0.98& & \\ \hline
\cite{aslam2023phish}(2023) & Mendeley & Open dataset~\cite{mendeley-url-data}& MLP, RF, RT, KNN, SVM & RF & 0.98&0.98&0.98& & \\ \hline
\cite{pathak2023classification}(2023) & UNK*& Previous work~\cite{pathak2023classification}& DT, KNN, SVM, NB, LR, XBoost, Aboost & Hybrid (DT, SVM, LR)  & 0.99&0.98 &0.99&0.99& \\ \hline
\cite{jishnu2023enhanced}(2023) & \url{phishTank}, Kaggle, and \href{https://majestic.com/reports/majestic-million}{Majestic}*& UNK &BERT &  BERT& 0.97&0.96&0.97&0.97& \\ \hline
\cite{rafsanjani2023qsecr}(2023) & \href{https://urlhaus.abuse.ch/}{URLHaus} and \url{phishtank.org} & Custom crawler & Custom rule based & Custom rule based & 0.93&0.93&0.93&0.93& \\ \hline
\cite{kalabarige2023boosting}(2023) & UCI ML Reposiroty and Mendeley & Open dataset~\cite{mendeley-url-data,UCI-ml-phishing-url,mendeley-url-phishing}& LGBM, XGBoost, AdaBoost, CatBoost, GB, Hybrid (BMLSELM) & Hybrid (BMLSELM) &0.97&0.97&0.97&0.97&   \\ \hline
\cite{ashwitha2023perception}(2023) & Kaggle* &UNK & LR, NB, DT, SVM, RF, KNN & KNN & & & 0.99& & \\ \hline
\cite{dr2023malicious}(2023) & Kaggle* &UNK& RF, XGBoost, LightGBBM &  RF&0.99&0.94&0.96&0.96&  \\ \hline
\cite{kundra2023identification}(2023) &UNK*& UNK & RF, AdaBoost, XGBoost, GBoost, KNN  & RF & & & 0.91& & \\ \hline
\cite{jha2023intelligent}(2023) & \url{phishtank.org}* & UNK&LR, RF & RF & 0.93&0.79&0.96&0.85& \\ \hline
\cite{nagy2023phishing}(2023) & PubMed & Open dataset~\cite{singh-url-phishing}&RF, NB, LSTM, CNN & UNK & & & & & \\ \hline
\cite{adebowale2023intelligent}(2023) & \url{phishtank.org} and \url{who.is}* & Custom crawler & LSTM, CNN, Hybrid (LSTM and CNN) & Hybrid (LSTM and CNN) & & &0.93& & \\ \hline
\cite{yu2024efficient}(2024) & Zhejiang Mobile Innovation Research Institute* & UNK& MBERT, XGBoost, LBoost, LSTM, NB, LR, RF, SVM, KNN & MBERT & 0.94&0.94& &094& \\ \hline
\caption{Data sources and Detection Methods used for Phishing URL detection. A single asterisk (*) indicates that the data is not publicly available. \emph{UNK} indicates \emph{Unclear/Unknown/Unspecified} details. Empty cells indicate missing values. \emph{P: Precision, R: Recall, A: Accuracy, F1: F1 Score, AUC: Area under the Curve}.}
\label{tbl:data_phishing_URL}
\end{longtable}
}
\vspace{-20pt}

{\footnotesize
\begin{longtable}{|l|p{4cm}|p{2.8cm}|p{3.3cm}|p{2.5cm}|l|l|l|l|l|}
\hline
\multirow{2}{*}{\textbf{\#}} & \multirow{2}{*}{\textbf{URL Source}} & \multirow{2}{*}{\textbf{Collection Method}} & \multirow{2}{*}{\textbf{Models Used}} & \multirow{2}{*}{\textbf{Best Model}} & \multicolumn{5}{|c|}{\textbf{Performance}}\\ 
\cline{6-10}
& & & & & \textbf{P} & \textbf{R} & \textbf{A} & \textbf{F1} & \textbf{AUC} \\ 
\hline
~\cite{salihovic2019role}(2019) &UNK* & UNK& RF, KNN, ANN, SVM, LR, NB&RF& & &0.97& & \\\hline
\cite{gallo2019identifying}(2019) & Spam emails received by a company* & UNK&GNB, DT, SVM, NN, RF & RF, SVM & 0.92&0.97&0.89& &\\ \hline
\cite{markova2019classification}(2019) & \url{cs.cmu.edu} & Open dataset~\cite{enron-email-dataset}& RF, KNN, SVM, DT &  RF&0.92&0.94&0.91& & \\ \hline
\cite{al2020email}(2020) & \url{aclweb.org} and previous work & Open dataset~\cite{acl-email-dataset} and previous work~\cite{almeida2011contributions}& NB, Dt, RF, SVM & SVM & 0.98&0.97&0.98&0.97& \\ \hline
\cite{rahmad2020performance}(2020) & Spam emails received by a company* & UNK & Clustering &Clustering & & &0.89& & \\ \hline
\cite{islam2021spam}(2021) & Open datasets* &UNK& LR, SVM, RF, XGBoost & XGBoost & & & & & \\ \hline
\cite{bhatti2021email}(2021) & \url{cs.cmu.edu}  &Open dataset~\cite{enron-email-dataset}& LSTM & LSTM & & &0.97& & \\ \hline
\cite{stojnic2021phishing}(2021) & UNK&UNK & Various topic modelling & N/A & N/A &N/A &N/A &N/A &N/A \\ \hline
\cite{genc2021understanding}(2021) & Author's spam folder* & Custom& LDA, Jaccard & N/A&N/A & N/A&N/A&N/A&N/A  \\ \hline
\cite{jonker2021using}(2021) & UNK* &UNK& RNN, LSTM, CNN, BERT & UNK& & & & & \\ \hline
\cite{janez2021trustworthiness}(2021) & Questionnaires and \url{untroubled.org/spam} & Open dataset\footnote{\url{https://untroubled.org/spam/}}&NB, SVM, RF, LR & NB & & &0.88&0.8& \\ \hline
\cite{venugopal2022detection}(2022) & UNK* & Custom crawler& BOW (Rule based) &BOW (Rule based)  & & &0.99& &\% \\ \hline
\cite{singh2022spam}(2022) & Kaggle&Open dataset~\cite{kagge-email-dataset} &CBoost, LR, DT, RF, GNB, SVM, KNN, XGBoost, LGBM, AdaBoost &CBoost & 0.97& &0.96&0.97& \\ \hline
\cite{livara2022empirical}(2022) & Kaggle&Open dataset~\cite{kagge-email-dataset2} & RF, NB, SVM, AdaBoost, LR & RF& & &0.99& & \\ \hline
\cite{al2022digital}(2022) & Previous work*  & Previous work~\cite{hina2021sefaced}& RF, LR, SVM, MNB & RF, LR, SVM, & 0.95&0.95&0.95& & \\ \hline
\cite{saka2022context}(2022) & GitHub, \href{https://monkey.org/~jose/phishing/}{monkey.org}, \url{cs.cmu.edu} & Open datasets~\cite{github-email-dataset,jose-monkey-email-dataset,enron-email-dataset}& K-Means, DBSCAN, and Agglomerative Clustering  & Agglomerative Clustering & N/A& N/A&N/A&N/A& \\ \hline
\cite{mughaid2022intelligent}(2022) & UNK & UNK&SVM, DT, LR, DNN, RF & DT & 1&1&1&1& \\ \hline
\cite{ismail2022efficient}(2022) & UCI ML Reposiroty*& UNK & NB, SVM, KNN, J48, DT  & DT & & &0.98& &  \\ \hline
\cite{janez2023review}(2023) & Previous work, \url{cs.cmu.edu}, \url{spamassassin.apache.org}, and \url{csmining.org}\footnote{Data link broken} & Previous work~\cite{androutsopoulos2000learning,cormack2007spam}, open datasets~\cite{enron-email-dataset,spamassasin-email-dataset}, and UNK& NB, SVM & UNK & & & & &  \\ \hline
\cite{mehdi2023adversarial}(2023) & Synthetic data & Data generated using various techniques~\cite{github-generated-email-dataset}& ALBERT, RoBERTa, BERT, DBERT, SQ, YOSO & ALBERT & & & 0.94&0.95& \\ \hline
\cite{ramprasath2023identification}(2023) & Kaggle* & UNK&RNN, LSTM, CNN & RNN & 0.99&0.92&0.99&0.95& \\ \hline
\cite{kushwaha2023analysis}(2023) & UCI ML Repository &Open dataset\cite{uci-sms-dataset} & BERT & BERT &0.95&0.93&0.98&0.94&  \\ \hline
\cite{saini2023machine}(2023) & UCI ML Repository* &Previous work~\cite{saini2023machine} & SVM, RF, NB &RF  & & &0.95& & \\ \hline
\cite{jena2023malicious}(2023) & Previous work* & Previous work~\cite{yerima2022semi}&KNN, NB, DT, RF, SVM, LR, XGBoost, BERT & BERT & 0.97&0.97&0.97&0.97& \\ \hline
\cite{mittal2023blockage}(2023) & UCI ML Repository*&UNK & CatBoost & CatBoost & 0.97&0.96&0.96&0.97&0.99 \\ \hline
\cite{bera2023towards}(2023) & Previous work, Kaggle, and \url{monkey.org} & Previous work~\cite{el2020depth,sakkis2003memory,metsis2006spam} and open datasets~\cite{jose-monkey-email-dataset,kaggle-email-dataset5}& Various topic modelling & N/A &N/A&N/A&N/A&N/A&N/A  \\ \hline
\cite{emmanuel2023information}(2023) & Kaggle&Open dataset~\cite{kagge-email-dataset3} & MLP, DT, LR, RF, KNN, SVM & MLP, SVM & &0.99&0.99&0.99&0.99   \\ \hline
\cite{chataut2024can}(2024) & Kaggle&Open dataset~\cite{kaggle-email-dataset4} & GPT-3.5, GPT-4, Custom (CyberGPT) &Custom (CyberGPT)  & & &0.97& & \\ \hline
\cite{jiang2024detecting}(2024) & UNK* & UNK& GPT-3.5, GPT-4 & UNK & & & & &  \\ \hline
\caption{Data sources and Detection Methods used for Phishing Email detection. A single asterisk (*) indicates that the data is not publicly available. \emph{UNK} indicates \emph{Unclear details}. Empty cells indicate missing values. \emph{P: Precision, R: Recall, A: Accuracy, F1: F1 Score, AUC: Area under the Curve}.}
\label{tbl:data_phishing_emails}
\end{longtable}
}
\vspace{-20pt}

{\footnotesize
\begin{longtable}{|l|p{4.5cm}|p{2.8cm}|p{3.2cm}|p{2.3cm}|l|l|l|l|l|}
\hline
\multirow{2}{*}{\textbf{\#}} & \multirow{2}{*}{\textbf{URL Source}} & \multirow{2}{*}{\textbf{Collection Method}} & \multirow{2}{*}{\textbf{Models Used}} & \multirow{2}{*}{\textbf{Best Model}} & \multicolumn{5}{|c|}{\textbf{Performance}}\\ 
\cline{6-10}
& & & & & \textbf{P} & \textbf{R} & \textbf{A} & \textbf{F1} & \textbf{AUC} \\ 
\hline
\cite{jain2019feature}(2019) & Previous work* & Previous work~\cite{almeida2011contributions} & SVM, LR, NN, NB, RF & RF& & &0.98& & \\\hline
\cite{zhang2020lies}(2020) & \href{https://ti.360.cn/}{360 Mobile Safe}*&UNK & SVM, NB, LR, RF &SVM  & 0.96&0.96& &0.96& \\\hline
\cite{liu2021detecting}(2021) & \href{https://ti.360.cn/}{360 Mobile Safe}* & UNK&LR, DT, NB, SVM & LR & 0.93&0.93& &0.93& \\\hline
\cite{ulfath2021hybrid}(2021)& UCI ML repository& Open dataset~\cite{uci-sms-dataset} & CNN, GRU, MLP, SVM, XGBoost, Hybrid (CNN, GRU) &Hybrid (CNN, GRU)& 0.99&0.96 & &0.98& \\\hline	
\cite{lai2022semorph}(2022) & \url{https://www.datafountain.cn/}*&Custom crawler& CNN, BERT, RoBERTa, ChineseBERT& Hybrid (Semorph/UNK) &0.96&0.84& &0.89& \\\hline
\cite{tang2022clues}(2022) & Twitter* & Twitter API & Custom/UNK & Custom/UNK&0.98&0.97&0.97& & \\\hline
\cite{jain2022sms}(2022) & UCI ML repository* & UNK&SVM, NB, LR, DT & SVM &0.96&0.93&0.98&0.95&  \\\hline
\cite{vinothkumar2022detection}(2022) & Kaggle and YouTube* &UNK& NB, DT, KNN & NB &  & &0.97& &\\\hline
\cite{abid2022spam}(2022) & Kaggle & Open dataset~\cite{kaggle-sms-dataset}&LR, SVC, RF, NB, GBM & RF &0.99&0.95&0.99&0.97& \\\hline
\cite{timko2023commercial}(2023) & \url{smishtank.com}*&Custom crawler & Various NLP methods &N/A  & & & & &  \\\hline
\cite{addanki2023safeguarding}(2023) & Kaggle and inaccessible website & Open dataset~\cite{kaggle-sms-dataset} and custom crawler&LinearDA, QDA, SVM, PCA, NB & SVM & & &0.97& & \\\hline
\cite{dharani2023spam}(2023) & Kaggle & Open dataset~\cite{kaggle-sms-dataset}&KNN, NB, RF, SVC, ETC, LR, XGBoost, AdaBoost, GBDT, DT, & NBB & 1& &0.95& & \\\hline
\cite{zhang2023bert}(2023) & Previous work* & Previous work~\cite{zhang2020lies}&BERT-GCN & BERT-GCN & &0.92&0.96&0.93& \\\hline
\cite{al2023multi}(2023) & UCI ML repository* &UNK &LSTM, CNN, RF, Hybrid (various), BERT, LSTM, XGBoost & Hybrid (CNN, LSTM) & 0.99&0.99&0.99&0.99& \\\hline
\cite{gandhi2023sms}(2023) & Kaggle &Open dataset~\cite{kaggle-sms-dataset}& DNN, LSTM &  DNN& & &0.95& & \\\hline
\cite{siagian2023improving}(2023) & UCI ML repository & Open dataset~\cite{uci-sms-dataset}&LSTM, GRU, NB, BERT & BERT & 0.99& &0.99& & \\\hline
\cite{kohilan2023machine}(2023) & Kaggle and Mendeley* & UNK&SNN, RNN, CNN & CNN & & & 0.99& & \\\hline
\cite{mishra2023dsmishsms}(2023) & UNK&UNK&BPA, RF, NB, DT & BPA &  & &0.97& &0.98 \\\hline
\cite{agrawal2023effective}(2023) & UNK &UNK& NB, RF, ETC &ETC  &0.99& &0.96& & \\\hline
\cite{seo2024device}(2024) & Korean Internet and Security Agency &UNK&NB, RF, LGBM, CNN, KoELECTRA & CharCNN & & &0.99&0.99& \\\hline
\caption{Data sources and Detection Methods used for Phishing SMS detection. A single asterisk (*) indicates that the data is not publicly available. \emph{UNK} indicates \emph{Unclear details}. Empty cells indicate missing values. \emph{P: Precision, R: Recall, A: Accuracy, F1: F1 Score, AUC: Area under the Curve}.}
\label{tbl:data_smishing}
\end{longtable}
}
\vspace{-20pt}

{\footnotesize
\begin{longtable}{|l|p{4.3cm}|p{3cm}|p{3.3cm}|p{2.5cm}|l|l|l|p{0.7cm}|p{0.7cm}|}
\hline
\multirow{2}{*}{\textbf{\#}} & \multirow{2}{*}{\textbf{Data}} & \multirow{2}{*}{\textbf{Collection Method}} & \multirow{2}{*}{\textbf{Models Used}} & \multirow{2}{*}{\textbf{Best Model}} & \multicolumn{5}{|c|}{\textbf{Performance}}\\ 
\cline{6-10}
& & & & & \textbf{P} & \textbf{R} & \textbf{A} & \textbf{F1} & \textbf{AUC} \\ 
\hline
\cite{ali2019detect}(2019) & Twitter user profiles & Open dataset~\cite{yang2013empirical} & GCNN, MLP, BP & GCNN & & & & &0.94 \\\hline
\cite{albayati2019identifying}(2019) & Facebook user profiles* & UNK & ID3, KNN, SVM & ID3 & 0.98&0.98&0.97& & \\\hline
\cite{yue2019madafe}(2019) & Twitter User Profiles & Open dataset~\cite{WU2018265} & SVM, RF, MADAFE (NN and LR) & MADAFE & & & & & \\\hline
\cite{venkatesan2019graph}(2019) & Twitter and Facebook (UNK)* & UNK & HDBSCAN & HDBSCAN & N/A &N/A&N/A&N/A&N/A \\\hline
\cite{raj2020multi}(2020) & Tweets* & Twitter API & KNN, RF, NB, DT & RF & & & & &0.95 \\\hline
\cite{zhang2021social}(2021) & Sina Weibo User profiles* & Custom crawler & CatBoost, RF & CatBoost & & & & &0.87 \\\hline
\cite{bebensee2021leveraging}(2021)  & Twitter user profiles & Open dataset~\cite{Cresci2018} & NB, QDA, SVM, KNN, RF, NN & RF & & &0.87&0.88&0.94 \\\hline
\cite{anklesaria2021survey}(2021)  & Instagram user profiles* & Instagram API & RF, AdaBoost, MLP, ANN, SGD & RF & 0.99&0.98&0.98&0.98&0.98 \\\hline
\cite{shreya2022identification}(2022)  & Facebook user profiles* & Manual collection & ANN, SVM, RF & ANN & & &0.96& & \\\hline
\cite{das2022effecient}(2022) & Instagram user profiles* & UNK & LR, KNN, SVM, RF, NB & RF & 0.99&0.97&0.94&0.98& \\\hline
\cite{rovito2022evolutionary}(2022) & Twitter user profiles & Open dataset~\cite{feng2021twibot} & GA, GP & GP &  & &0.76&0.78& \\\hline
\cite{shukla2022tweezbot}(2022) & Twitter user profiles & Open dataset~\cite{jain_kaggle} & SVM, CNB, BNB, MP, DT, RF & TweezBot (Unclear) & 0.99&0.93&0.98& &0.99 \\\hline
\cite{na2023evolving}(2023) & YouTube video and user metadata and comments* & YouTube API & Sentence-BERT, RoBERTa, YouTuBERT & YouTuBERT (LLM and DBSCAN) & 0.63 & &0.81 &0.90& 0.71 \\\hline
\cite{haq2023spammy}(2023) & List of names* & UNK & NB, KNN, SVM, LR, RF & NB & 0.94& 0.94 & 0.95& 0.94& \\\hline
\cite{gangan2023detection}(2023) & Twitter user profiles* & UNK & NB, DT, NN, Ensemstack & Ensemstack & & &0.98& & \\\hline
\cite{fathima2023ann}(2023) & Instagram user profiles* & UNK & ANN & ANN & & & & 0.74& \\\hline
\cite{nikhitha2023fake}(2023) & Instagram user profiles* & UNK &  LR, DT, RF & RF & & & 0.9 & & \\\hline
\cite{singh2023safe}(2023) & Twitter user profiles and tweets* & Twitter API & LR & LR & 0.93&0.93&0.93&0.93& \\\hline
\caption{Data sources and Detection Methods used for Fake User detection. A single asterisk (*) indicates that the data is not publicly available. \emph{UNK} indicates \emph{Unclear details}. Empty cells indicate missing values. \emph{P: Precision, R: Recall, A: Accuracy, F1: F1 Score, AUC: Area under the Curve}.}
\label{tbl:data_fake_users}
\end{longtable}
}
\vspace{-20pt}

{\footnotesize
\begin{longtable}{|l|p{4cm}|p{3.5cm}|p{3cm}|p{2.5cm}|l|l|l|p{0.7cm}|p{0.7cm}|}
\hline
\multirow{2}{*}{\textbf{\#}} & \multirow{2}{*}{\textbf{Job postings source}} & \multirow{2}{*}{\textbf{Collection Method}} & \multirow{2}{*}{\textbf{Models Used}} & \multirow{2}{*}{\textbf{Best Model}} & \multicolumn{5}{|c|}{\textbf{Performance}}\\ 
\cline{6-10}
& & & & & \textbf{P} & \textbf{R} & \textbf{A} & \textbf{F1} & \textbf{AUC} \\ 
\hline
\cite{lal2019orfdetector}(2019) & Kaggle & Open dataset~\cite{emscad-dataset} & J48, LR, RF & Ensemble & & &0.95&0.94& \\\hline
\cite{ranparia2020fake}(2020) & Kaggle & Open dataset~\cite{emscad-dataset} & GLoVE & GLoVE & & &0.99& & \\\hline
\cite{nasser2021online}(2021) & Kaggle & Open dataset~\cite{emscad-dataset} & ANN & ANN & 0.91&0.96& &0.93& \\\hline
\cite{vo2021dealing}(2021) & Kaggle & Open dataset~\cite{emscad-dataset} & LR & LR & &0.89&0.92& &0.96 \\\hline
\cite{li2021exploratory}(2021) & Kaggle & Open dataset~\cite{emscad-dataset}& LightGBM, LR, DT, XGBoost, AdaBoost & LightGBM &0.93&0.94&0.95&0.93& \\\hline
\cite{tabassum2021detecting}(2021) & job.com.bd, bdjobstoday, and deshijob* & Custom crawler & LR, AdaBoost, DT, RF, VC, LGBM, GB & LightGBM or GBoost & & &0.95& & \\\hline
\cite{habiba2021comparative}(2021) & Kaggle & Open dataset~\cite{emscad-dataset} & KNN, RF, DT, SVM, NB, DNN &  DNN & & &0.97& & \\\hline
\cite{amaar2022detection}(2022) & Kaggle & Open dataset~\cite{emscad-dataset} & RF, LR, SVM, ETC, KNN, MP & ETC & & &0.99& & \\\hline
\cite{bhatia2022detection}(2022) & Kaggle & Open dataset~\cite{emscad-dataset} & KNN, RF & KNN & 0.79&0.73&0.98&0.76& \\\hline
\cite{mahbub2022online}(2022)	& SEEK, Glassdoor, Indeed, and Gumtree job postings* & Custom crawler & RF, JRip, NB, J48 & RF & 0.82&0.69&0.91& & \\ \hline
\cite{nessa2022recruitment}(2022) & Kaggle & Open dataset~\cite{emscad-dataset} & GRU & GRU & & & &0.93& \\\hline
\cite{prathaban2022verification}(2022) & Kaggle & Open dataset~\cite{emscad-dataset} & LR, NB, MLP, KNN, RF, DT, Adaboost, GB, NLP & RF & 0.98&0.97&0.97&0.98& \\\hline
\cite{pandey2022effective}(2022) & Kaggle & Open dataset~\cite{emscad-dataset} & RF, SVM, Bi-LSTM & Bi-LSTM & & &0.98&0.98& \\\hline
\cite{yang2023improved}(2023) & Kaggle & Open dataset~\cite{emscad-dataset} & SVM, NB, RF, Bi-LSTM, LR & RF & & & & & \\\hline
\cite{reddy2023web}(2023)  & Kaggle & Open dataset~\cite{emscad-dataset} & RF, XBoost, LightGBM, CatBoost, DT & XGBoost & 0.95&0.9&0.96&0.92& \\\hline
\cite{santhiya2023fake}(2023)  & Kaggle & Open dataset~\cite{emscad-dataset} & RF, SVM, NB, Ensemble & RF & & &0.98& & \\\hline
\cite{sofy2023intelligent}(2023)  & Kaggle & Open dataset~\cite{emscad-dataset} & RF, NB, SVM, DT, KNN & RF & & &0.97& & \\\hline
\cite{nanath2023investigation}(2023)  & Kaggle & Open dataset~\cite{emscad-dataset} & LR, DT, RF, NB, GLM &  GLM & 0.96&0.78& &0.86&0.98\\\hline
\cite{ullah2023smart}(2023)  & Kaggle & Open dataset~\cite{emscad-dataset} & AdaBoost, XGBoost, RF, Voting & AdaBoost &0.99&0.97&0.98&0.98& \\\hline
\cite{zhang2023orfpprediction}(2023)  & Boss Zhipin, Liepin, 51job*& Custom crawler& NB, XGBoost, SVM, LightGBM, DT, RF & DRLM (DT and RF and LightGBM) & &0.98&0.94&0.92& \\\hline
\caption{Data sources and Detection Methods used for Fraudulent Recruitment detection. The asterisk (*) indicates that the data is not publicly available. Empty cells indicate missing values. \emph{P: Precision, R: Recall, A: Accuracy, F1: F1-Score, AUC: Area Under the Curve}.}
\label{tbl:data_recruitment}
\end{longtable}} 
\vspace{-20pt}

{\footnotesize
\begin{longtable}{|l|l|p{3.5cm}|p{3cm}|p{2.5cm}|l|l|l|p{0.7cm}|p{0.7cm}|}
\hline
\multirow{2}{*}{\textbf{\#}} & \multirow{2}{*}{\textbf{Data}} & \multirow{2}{*}{\textbf{Collection Method}} & \multirow{2}{*}{\textbf{Models Used}} & \multirow{2}{*}{\textbf{Best Model}} & \multicolumn{5}{|c|}{\textbf{Performance}}\\ 
\cline{6-10}
& & & & & \textbf{P} & \textbf{R} & \textbf{A} & \textbf{F1} & \textbf{AUC} \\ 
\hline
\cite{gupta2020detecting}(2020)  & Amazon reviews* & Custom crawler & SVM, LR, RF, DT, GNBSGD, KNN, 3LP, 4LP, XGBoost & 3LP & 0.98 & 0.98 & & 0.98 & 0.98 \\\hline
\cite{furia2020tool}(2020) & Amazon reviews* & Custom crawler & SVM, KNN, NB, Ensemble & Ensemble & 0.81 & 0.81 & 0.81 & 0.81 & \\\hline
\cite{wang2021modeling}(2021) & Yelp and JD.com reviews & Open dataset\cite{yelpzip,yelpnyc} and JD.com custom crawler & GraphSAGE, Cluster-GCN, HGT, \emph{C-FATH (Custom)} & C-FATH (Unclear)* & & & & 0.68-0.87 & 0.95-0.97 \\ \hline
\cite{chandana2021analyzing}(2021) & Amazon reviews & Open dataset~\cite{liu2019method} & RF & RF* & 1 & 0.85 & 0.98 & & \\\hline
\cite{javed2021fake}(2021) & Yelp reviews* & UNK & CNN, SVM, LR, MLP & CNN & 0.93 & 0.92 & 0.92 & & \\\hline 
\cite{balakrishna2022identifying}(2022) & Yelp reviews & Open dataset~\cite{asghar2016yelp,ott2013negative} & WaveNet, LDA & WaveNet, LDA & N/A& N/A& N/A& N/A& N/A \\\hline
\cite{deekshan2022detection}(2022) & Amazon reviews* & UNK & BERT, VADER, LSTM, WordNet, SGD, SVM, LR & LR & & & 0.81 & & \\\hline
\cite{rangari2022empirical}(2022) & Amazon hotel reviews* & UNK & KNN, NB, SVM & SVM* & & & 0.93 & & \\\hline
\cite{tushev2022domain}(2022) & Smartphone App reviews* & Web Scraping & LDA, keyATM & keyATM & N/A & N/A& N/A&N/A & N/A\\ \hline
\cite{obie2022violation}(2022) & Smartphone App reviews & Open dataset~\cite{eler2019android,obie2021first} & SVM, DT, NN, LR, GBT & SVM & 0.94 & 0.84 & & 0.89 & \\ \hline
\cite{yugeshwaran2022rank}(2022) & Google Play reviews* & Custom crawler & DT, RF, MLP & MLP & & & 0.97 & & \\\hline
\cite{rangar2022machine}(2022) & Amazon reviews* & UNK & CNN, SVM, NB & CNN & 1 & 1 & & 1 & \\\hline
\cite{tufail2022effect}(2022) & Hotel reviews & Open dataset~\cite{ott2013negative,yelp-dataset} & SVM, KNN, LR & SKL (SVM and KNN and LR) & & & 0.95 & & \\\hline
\cite{harris2022combining}(2022) & Yelp reviews & Open dataset~\cite{yelpzip} & Bi-LSTM & Bi-LSTM & & & & & 0.89 \\\hline
\cite{akshara2023small}(2023) & Amazon book reviews* & UNK & SVM, LR & LR & & & 0.86 & & \\\hline
\cite{singh2023deep}(2023) & Reviews & Open dataset~\cite{ott2013negative} and UNK & CNN, LSTM, KNN, NB, SVM, W2V & CNN, LSTM & & & 0.93 & & \\\hline
\cite{iqbal2023efficient}(2023) & Amazon reviews* & Custom crawler & AdaBoost, RF, Lr, SVM, KNN & RF & 0.99 & 0.99 & 0.99 & 0.99 & \\\hline
\cite{ashraf2023fake}(2023) & Yelp reviews* & UNK & SVM, MLP, CNN, LR & CNN & 0.85 & 0.85 & 0.85 & 0.85 & \\\hline
\cite{silpa2023detection}(2023) & Yelp reviews* & UNK & NB, LR, SVM, DT & SVM & 0.96 & 0.98 & 0.97 & & \\\hline
\cite{pengqi2023unmasking}(2023) & Yelp reviews & Open dataset~\cite{yelpzip} & GPT-3, BERT, RF, XGBoost & GPT-3 & 0.73 & 0.64 & & 0.68 & 0.75 \\\hline
\cite{ganesh2023implementation}(2023) & Undefined reviews* & UNK & ANN, CNN, LR, SVM, NB, KNN, RF, DT, SGD & LR & & & 0.89 & & \\\hline
\cite{thilagavathy2023fake}(2023) & Hotel reviews* & UNK & SVM & SVM & & & & &  \\\hline
\cite{bevendorff2024product}(2024) & Product reviews* & YouTube API & SVM,LR & LR & 0.74 & 0.99 & & 0.85 & 0.95\\ \hline
\caption{Data sources and Detection Methods used for Fake Review detection. A single asterisk (*) indicates that the data is not publicly available. \emph{UNK} indicates \emph{Unclear details}. Empty cells indicate missing values. \emph{P: Precision, R: Recall, A: Accuracy, F1: F1-Score, AUC: Area Under the Curve}.}
\label{tbl:data_fake_reviews}
\end{longtable}}
\end{landscape}
\twocolumn

%%%%%%%%%%%%%%%%%%%%%%%%%%%%%%%%%%%%%%%%%%%%%%
%%                                          %%
%% Backmatter begins here                   %%
%%                                          %%
%%%%%%%%%%%%%%%%%%%%%%%%%%%%%%%%%%%%%%%%%%%%%%

\begin{backmatter}

\section*{Acknowledgements}%% if any
The authors would like to thank the reviewers for their feedback and reviews.

\section*{Funding}%% if any
% This section is removed for the review process.
This work was funded by The Dawes Trust.

\section*{Abbreviations}%% if any
PRISMA-ScR: Preferred Reporting Items for Systematic reviews and Meta-Analyses extension for Scoping Reviews; NLP: Natural Language Processing; AI: Artificial Intelligence; ML: Machine Learning; SLR: Systematic Literature Review.

\section*{Availability of data and materials}%% if any
The data extracted from the academic papers included in this review can be found in this public repository: \url{https://osf.io/nrx7y/?view_only=ca1050d48c4c4a969817c6d5f677cb87}.

\section*{Ethics approval and consent to participate}%% if any
Not applicable.

\section*{Competing interests}
The authors declare that they have no competing interests.

\section*{Consent for publication}%% if any
Not applicable.

\section*{Authors' contributions}
% This section is removed for the review process.
AP and NT drafted the final manuscript. 
AP conducted the updated academic literature review, designed and conducted the grey literature review, extracted data from the literature, and analysed and interpreted the results of all aspects of the scoping review. 
NT contributed to the coding process, and supervised all aspects of the study.
SJ and ED provided substantial feedback on the review process and editing of the document. 
EM contributed to the coding process.
AM and SL contributed to the editing process. 

\section*{Authors' information}%% if any
% This section is removed for the review process.
\textsuperscript{1} Security and Crime Science, University College London, London, United Kingdom.\\
\textsuperscript{2} Humanities and Social Sciences, Anglia Ruskin University, London, United Kingdom.\\
\textsuperscript{3} Advanced Research Computing Centre, University College London, London, United Kingdom.

%%  After submission of the .TEX file,                     %%
%%  you will be prompted to submit your .BBL file.         %%

% if your bibliography is in bibtex format, use those commands:
\bibliographystyle{bmc-mathphys} % Style BST file (bmc-mathphys, vancouver, spbasic).
\bibliography{references}      % Bibliography file (usually '*.bib' )
% for author-year bibliography (bmc-mathphys or spbasic)
% a) write to bib file (bmc-mathphys only)
% @settings{label, options="nameyear"}

% or include bibliography directly:
% \begin{thebibliography}
% \bibitem{b1}
% \end{thebibliography}

%%%%%%%%%%%%%%%%%%%%%%%%%%%%%%%%%%%
%%                               %%
%% Figures                       %%
%%                               %%
%% NB: this is for captions and  %%
%% Titles. All graphics must be  %%
%% submitted separately and NOT  %%
%% included in the Tex document  %%
%%                               %%
%%%%%%%%%%%%%%%%%%%%%%%%%%%%%%%%%%%

%%%%%%%%%%%%%%%%%%%%%%%%%%%%%%%%%%%
%%                               %%
%% Additional Files              %%
%%                               %%
%%%%%%%%%%%%%%%%%%%%%%%%%%%%%%%%%%%

\end{backmatter}
\end{document}